\documentclass{article}

\PassOptionsToPackage{numbers, compress}{natbib}

\usepackage[preprint]{neurips_2026}
\usepackage[utf8]{inputenc} 
\usepackage[T1]{fontenc}    
\usepackage{hyperref}       
\usepackage{url}            
\usepackage{booktabs}       
\usepackage{amsfonts}       
\usepackage{nicefrac}       
\usepackage{microtype}      
\usepackage{multicol}  
\usepackage{enumitem}
\usepackage{bbding}
\usepackage{booktabs}
\usepackage{url}
\usepackage{bm}  

\def\x{\mathbf{x}}
\def\Phi{\boldsymbol{\phi}}

\usepackage[table]{xcolor}  
\usepackage{hyperref}

\usepackage[algo2e,linesnumbered,ruled,vlined]{algorithm2e}
\definecolor{cvprblue}{rgb}{0.21,0.49,0.74}
\usepackage{amsmath}
\usepackage{amssymb}
\usepackage{mathtools}
\usepackage{amsthm}
\usepackage{cuted}
\usepackage{float}
\usepackage{graphicx}
\usepackage{pifont}
\usepackage{threeparttable}
\usepackage{multirow}
\usepackage{microtype}
\definecolor{ourscore}{HTML}{F1F5FB}

\usepackage[capitalize,noabbrev]{cleveref}
\usepackage{wrapfig}
\theoremstyle{plain}
\newtheorem{theorem}{Theorem}[section]

\theoremstyle{definition}

\newtheorem{assumption}[theorem]{Assumption}
\theoremstyle{remark}

\usepackage[textsize=tiny]{todonotes}
\usepackage{caption}
\title{Anchoring Adversarial Trajectories to Data Manifolds: A Bilevel Transfer Optimization Framework}

\author{%
  Yaohua Liu\\
  The University of Hong Kong\\
  Hong Kong, China\\
  \texttt{liuyaohua.918@gmail.com}
  \And
  Yifan Guo\\
  International School of Information Science \& Engineering\\
  Dalian University of Technology, Dalian, China\\
  \texttt{friscoguo@gmail.com}
  \And
  Jiaxin Gao\thanks{Corresponding author.}\\
  The Hong Kong Polytechnic University\\
  Hong Kong, China\\
  \texttt{jiaxinm.gao@outlook.com}
}

\begin{document}

\maketitle

\begin{abstract}

			A key bottleneck in adversarial transfer is a \textit{trajectory-level geometric disconnect}: ambient gradients often drift away from the intrinsic data manifold, causing surrogate-specific overfitting. To rectify this, we propose \textit{Manifold Anchored Bilevel Transfer (MABT)}, a unified framework that anchors adversarial trajectories to the shared semantic subspace. MABT introduces a relaxed manifold-anchoring operator as a semantic rectifier to suppress off-manifold noise. With this constraint, we cast transfer attack generation as a distributional bilevel optimization problem that learns a geometry-aligned initialization by minimizing expected transfer risk under a surrogate uncertainty distribution. We further develop a Hessian-free solver with linear-time complexity to handle the resulting hierarchy. Experiments demonstrate improved transferability for \textbf{10} baseline attackers across \textbf{28} attack configurations, diverse victim architectures, and defense mechanisms.
\end{abstract}

		\begin{figure}[H]
	\centering
	\includegraphics[width=0.99\textwidth]{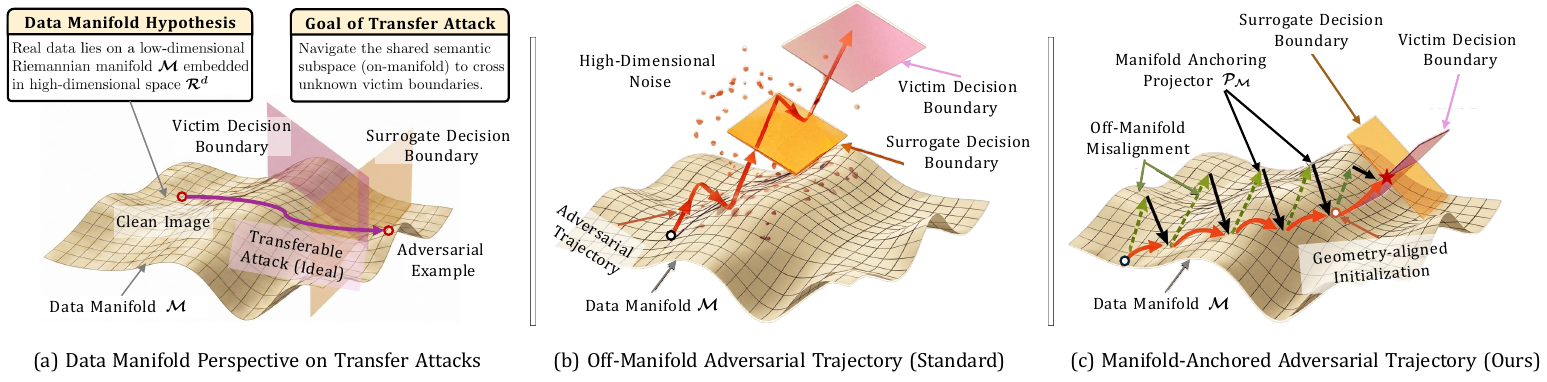}\\
	\caption{
		\textbf{Motivation.}
		(a) \textbf{Manifold hypothesis}: Natural images concentrate near a low-dimensional manifold $\mathcal{M}$ capturing intrinsic semantics, where decision boundaries of diverse models tend to partially align.
		(b) \textbf{Off-manifold divergence}: Standard iterative attacks inherently drift into high-dimensional noise away from $\mathcal{M}$; in this region, boundaries are less consistent, so the trajectory may cross the \textbf{\textcolor{orange}{surrogate boundary}} but miss the \textbf{\textcolor{purple}{victim boundary}}.
		(c) \textbf{Manifold anchoring}: MABT uses a manifold-anchoring projector with a geometry-aligned initialization to suppress off-manifold deviations while preserving task-relevant semantics to cross \textit{shared boundaries}.
	} 
	\label{fig:FirstFigure}
	\vspace{-0.3cm} %
\end{figure}

\section{Introduction}

Adversarial examples (AEs), crafted by adding imperceptible perturbations to clean inputs, expose the fragility of deep learning models~\cite{carlini2017towards,madry2017towards,yin2025adversarial}. In real-world scenarios where model internals are often inaccessible, \emph{transfer-based attacks} pose a critical security threat~\cite{papernot2016transferability,bai2025rat}. By generating AEs on a white-box surrogate that can mislead unknown black-box victims, these attacks exploit the transferability of adversarial features across different architectures~\cite{dong2018boosting,wu2024attacksadversarialmachinelearning}.

To boost transferability, extensive research has been devoted to improving attack generalization from multiple perspectives. 
Broadly, existing methods can be categorized into three primary streams: 
(i) \emph{Gradient Optimization}, which stabilizes update directions via momentum or variance reduction~\cite{dong2018boosting,lin2019nesterov,wang2021boosting}; 
(ii) \emph{Input Transformations}, which augment inputs to mitigate overfitting to the surrogate's decision boundary~\cite{xie2019improving,dong2019evading,long2022frequency}; 
and (iii) \emph{Model-Refinement and Ensemble Strategies}, which diversify surrogates by modifying architectures~\cite{11018095,liu2025harnessing} or aggregating multi-model gradients~\cite{li2020learning,li2023making}. 
In addition, recent studies have explored feature-level objectives~\cite{liupixel2feature,10833658} and generative priors~\cite{wu2025boosting}, though these strategies often involve heuristic data augmentations or model priors. 
Despite their success, these advances mainly refine isolated factors in the attack pipeline, leaving the underlying mechanism of transferability less explicitly characterized.

We revisit the motivation (Fig.~\ref{fig:FirstFigure}) from the geometry of the optimization process, and attribute the transferability bottleneck to a \emph{geometric misalignment} that unfolds in three stages. \textit{(i) Manifold premise.} Prior work in vision~\cite{li2025back} and classical manifold theory~\cite{chapelle2006discussion} suggest that natural images concentrate near a low-dimensional manifold embedded in the high-dimensional pixel space; within this semantically structured region, decision boundaries of different models are more likely to exhibit partial alignment. \textit{(ii) Off-manifold drift.} However, standard iterative attacks are driven by ambient-space gradients and, even under $\ell_p$ constraints, tend to accumulate perturbations dominated by components orthogonal to the manifold. Consequently, the adversarial trajectory gradually departs from the shared semantic subspace and enters off-manifold regions where the natural-image prior provides little geometric guidance. \textit{(iii) Misalignment.} Once this drift occurs, optimization increasingly overfits surrogate-specific artifacts; in these ambient regions, decision boundaries can differ substantially across architectures, which in turn weakens transferability.

\begin{wrapfigure}{r}{7cm}
	\vspace{-0.4cm} %
	\begin{minipage}[t]{0.5\textwidth}
		\centering
		\includegraphics[width=\linewidth]{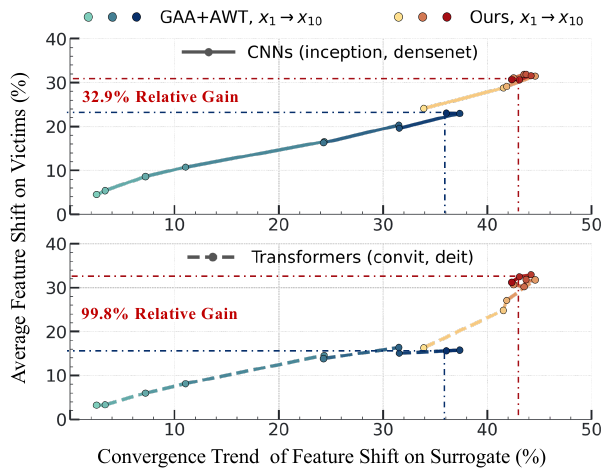}
\caption{\textbf{Feature space divergence.} We quantify victim-side feature shift during the attack process as $1-\cos(f_{clean},f_{adv})$, where $\x_1\!\rightarrow\!\x_{10}$ denotes the adversarial trajectory over 10 attack iterations. Compared with GAA~\cite{gan2025boosting}+AWT~\cite{chen2025enhancing}, MABT induces substantially larger semantic shifts on victim models, with relative gains of \textbf{99.8\%} on Transformers and \textbf{32.9\%} on CNNs.}
\label{fig:feature_shift_gain}
\label{fig:feature_shift_gain}
	\end{minipage}
	\vspace{-0.3cm}
\end{wrapfigure}

\subsection{Our Contributions}

This geometric perspective suggests that improving transferability requires more than strengthening a single attack component. Instead, the attack trajectory should be explicitly regularized so that its evolution remains aligned with semantically meaningful directions that are more likely to be shared across models. As illustrated in Fig.~\ref{fig:feature_shift_gain}, once such geometric consistency is better preserved, the induced feature shift on victim models grows more steadily rather than saturating early due to surrogate overfitting.

To bridge this geometric disconnect, we propose \textit{\textbf{M}anifold-\textbf{A}nchored \textbf{B}ilevel \textbf{T}ransfer (MABT)}, a unified framework that explicitly anchors adversarial trajectories to the intrinsic data geometry (Fig.~\ref{fig:FirstFigure}c).  First, we construct a relaxed manifold-anchoring operator that actively suppresses off-manifold components while preserving task-relevant semantics. Building on this geometric foundation, we formulate transfer attack as a distributional bilevel optimization problem, which learns a geometry-aligned initialization by minimizing the \emph{expected transfer risk} over a surrogate uncertainty distribution. To handle the resulting hierarchical structure efficiently, we further develop a Hessian-free solver based on the regularized gap function. Overall, by enforcing geometric consistency throughout optimization, MABT maintains a more stable positive correlation between surrogate-side and victim-side feature shifts, thereby inducing more universal adversarial features and consistently improving transferability across disparate architectures. We summarize the  main contributions  as follows:
\begin{itemize}
	
	\item 
	We revisit transfer attacks through a \emph{manifold anchoring} lens, showing that off-manifold drift is a critical bottleneck hindering transferability and motivating a unified framework to correct the resulting geometric misalignment.
	
	\item 
	We introduce the Manifold Anchoring Projector (MAP), a relaxed anchoring operator that filters surrogate-specific off-manifold noise, preserves task-relevant semantics, and supports flexible geometric instantiations.

	\item 
	We formulate transfer attacks as \emph{Distributional Bilevel Optimization (DBO)} to learn a geometry-aligned initialization by minimizing expected transfer risk. We further introduce a Hessian-free solver with linear complexity to ensure efficient convergence.

	\item 
	Extensive evaluation on 12 victim models spanning CNNs and Transformers demonstrates improved transferability across \textbf{10} baseline attackers and \textbf{28} attack configurations under diverse attack and defense settings.
\end{itemize}



%

\section{Related Work}
\textbf{Transfer-based Adversarial Attacks.} 
Current transfer attacks generally follow an iterative optimization paradigm and refine specific components within this loop. 
\textit{(1) Gradient and Input Refinement}. To stabilize update directions, Gradient-based methods introduce momentum terms~\cite{dong2018boosting,wang2021boosting}, Nesterov acceleration~\cite{lin2019nesterov}, or variance reduction techniques~\cite{wang2021enhancing} to escape local optima. 
Simultaneously, Input-based strategies augment the input space to mitigate overfitting, employing diverse transformations such as random resizing~\cite{xie2019improving}, translation~\cite{dong2019evading}, scale-invariance~\cite{lin2019nesterov}, and patch-wise mixing~\cite{wang2021admix,gu2022segpgd}. 
\textit{(2) Advanced Objective Design.} Beyond standard losses, recent works design specialized objectives to disrupt intrinsic representations. This includes targeting intermediate feature layers~\cite{ganeshan2019fda,huang2019enhancing}, fusing multi-scale attention maps~\cite{10833658}, or shifting the perturbation domain from pixels to features~\cite{liupixel2feature}. 
Besides, several works also leverage architecture-specific priors such as eroding feature responses in Ghost networks~\cite{li2020learning}, utilizing skip connections~\cite{wu2020boosting}, or harnessing the computational redundancy in ViTs~\cite{liu2025harnessing} and alignment modules~\cite{11018095}.

Another line of work improves transferability by modifying the surrogate side or learning generator-based attack mechanisms.
Ensemble-based attacks aggregate gradients from multiple models~\cite{liu2017delving,li2023making} or stochastic variants~\cite{cai2023ensemble} to reduce model bias.
Furthermore, generative approaches~\cite{Poursaeed_2018_CVPR,wu2025boosting} and meta-learning frameworks~\cite{yuan2021meta,fang2022learning} attempt to learn transferable perturbation priors or generalizable attack generators. Beyond adversarial transfer, diffusion models have also been used to model long-horizon trajectories in offline goal-conditioned reinforcement learning~\cite{zhang2026diffusionsubgoalplanninglonghorizon}.
However, these methods still improve transferability mainly through surrogate diversity or learned generation, without explicitly modeling how the adversarial trajectory evolves geometrically.
\textit{Summary}. In contrast to existing works that prioritize specific isolated factors, we investigate the optimization process from the perspective of the \emph{adversarial trajectory}. We propose a unified framework that explicitly aligns perturbation evolution with the data manifold, while enabling the initialization to learn transferable geometric priors to rectify the geometric disconnect.


\textbf{Manifold Perspectives in Deep Learning.}
Beyond specific attack tactics, the manifold hypothesis~\cite{chapelle2006discussion,li2025back} also serves as a geometric prior in broader visual tasks. 
In the security realm, this perspective has been adopted for defense mechanisms. 
Approaches utilize manifold manipulation to purify adversarial perturbations~\cite{zhou2020manifold} or detect anomalies that deviate from the natural distribution~\cite{NEURIPS2020_23937b42}. 
A recent theoretical study also analyzes transfer-based attacks from a manifold perspective, explaining transferability through the geometry and curvature of the data manifold~\cite{chen2023theory}.
On the offensive front, manifold priors are leveraged to craft stealthy or realistic examples. 
These methods generally constrain perturbations to bypass detection~\cite{Bar_Tal_2023_ICCV}, ensure geometric consistency in 3D domains~\cite{ghosh2025targeted}, or perform unrestricted attacks by optimizing adversarial directions in generative latent manifolds~\cite{chen2023content}. \textit{Summary}. While these studies highlight the relevance of manifold geometry, they mainly use it for purification, detection, theoretical explanation, or realistic example generation. Different from these approaches, our work uses manifold anchoring as a semantic rectifier to explicitly guide the optimization trajectory for improved transferability.

\textbf{Bilevel Optimization and Applications.} Bilevel optimization provides a hierarchical framework for coupled decision-making and has been widely applied in meta-learning~\cite{franceschi2018bilevel} and neural architecture search~\cite{liudarts}. Recent advances have gradually shifted from expensive Hessian-based methods to more efficient first-order solvers~\cite{liu2021value,yaoovercoming,ji2021bilevel}. In trustworthy machine learning, bilevel formulations are mainly used to optimize robustness-related objectives~\cite{madry2017towards, zhang2022revisiting}.  For transfer attacks, recent work~\cite{liu2024advancing} has explored bilevel initialization optimization, but it relies on auxiliary victim models to construct the objective and requires high-order gradient estimation, leading to extra dependencies and computational overhead. \textit{Summary}. In contrast, MABT formulates a distributional bilevel optimization problem that minimizes the \emph{expected transfer risk}. This design eliminates the need for extra model accesses and employs a Hessian-free solver, enabling efficient learning of geometry-aligned priors for improved transferability.


\section{Methodology}

\subsection{Perspective}
\label{sec:map_perspective}


Let $\mathcal{D}$ denote the data distribution supported on a low-dimensional Riemannian manifold $\mathcal{M} \subset \mathbb{R}^d$, embedded within the ambient pixel space $\mathbb{R}^d$. Given a clean pair $(\mathbf{x}, \mathbf{y}) \in \mathcal{M} \times \mathcal{Y}$ and a white-box surrogate $\mathcal{S}$, existing iterative attacks (e.g., PGD) typically generate perturbations by ascending surrogate gradients inside an $\ell_p$-norm ball $\Omega = \{\boldsymbol{\delta} \in \mathbb{R}^d \mid \|\boldsymbol{\delta}\|_p \le \epsilon\}$, without explicitly modeling the manifold geometry of natural images. The update rule is governed by:
\begin{equation}
	\boldsymbol{\delta}_{t+1} = \Pi_{\Omega} \left( \boldsymbol{\delta}_t + \alpha \cdot \text{sign}\left( \nabla_{\mathbf{x}} \mathcal{L}(\mathcal{S}(\mathbf{x} + \boldsymbol{\delta}_t), \mathbf{y}) \right) \right),
	\label{eq:ambient_update}
\end{equation}
where $\Pi_{\Omega}$ projects the update onto the feasible set, and $\text{sign}(\cdot)$ is applied element-wise.

\noindent\textbf{Off-manifold Geometric Misalignment.}
While Eq.~\eqref{eq:ambient_update} degrades $\mathcal{S}$ via high-frequency gradients, it suffers from intrinsic \textit{geometric misalignment}. Driven by ambient-space gradients, the accumulated perturbation $\boldsymbol{\delta}$ is dominated by orthogonal components ($\mathcal{M}^{\perp}$) that pull the trajectory away from $\mathcal{M}$. Consequently, the attack drifts into undefined ``off-manifold'' regions of surrogate-specific artifacts, which fail to transfer to unknown victims $\mathcal{V}$ due to the boundary divergence shown in Fig.~\ref{fig:FirstFigure}b.

\noindent\textbf{Manifold Anchoring Projector.}
To rectify this geometric disconnect, we move beyond standard $\ell_p$-bounded attacks defined purely in the ambient space and introduce a \textit{manifold-anchored} evolution process. We propose the MAP, a generalized operator $\mathcal{P}_{\mathcal{M}}: \mathbb{R}^d \to \mathbb{R}^d$ designed to \emph{anchor} ambient adversarial states toward the intrinsic data manifold by suppressing off-manifold components, while strictly preserving the perturbation budget.

We formulate $\mathcal{P}_{\mathcal{M}}$ as a \textit{generalized variational anchoring problem}. Given a query state $\mathbf{z} \in \mathbb{R}^d$ (e.g., $\mathbf{z} = \mathbf{x} + \boldsymbol{\delta}$), the ideal anchoring operator seeks an anchored state $\mathbf{u}^*$ by solving:
\begin{equation}
	\label{eq:map_variational}
	\mathcal{P}_{\mathcal{M}}[\mathbf{z}] :=
	\operatorname*{arg\,min}_{\mathbf{u} \in \mathbf{x}+\Omega} \quad
	\Big( \underbrace{\Psi(\mathbf{u}, \mathbf{z})}_{\text{Fidelity Potential}}
	+ \underbrace{\mathcal{R}_{\mathcal{M}}(\mathbf{u})}_{\text{Manifold Prior}} \Big),
\end{equation}
where $\Psi(\mathbf{u}, \mathbf{z})$ enforces signal fidelity (e.g., Euclidean or perceptual distance), and $\mathcal{R}_{\mathcal{M}}(\cdot)$ penalizes deviations from the data manifold. Here $\mathbf{x}$ is omitted from $\mathcal{P}_{\mathcal{M}}$ for notational simplicity. The feasible set $\mathbf{x}+\Omega$ enforces $\|\mathbf{u}-\mathbf{x}\|_p\le\epsilon$, ensuring strict compliance with the adversarial constraints.

We emphasize that Eq.~\eqref{eq:map_variational} provides a \textbf{generalized formulation of manifold anchoring}, and does not imply that every practical MAP instance exactly solves the variational problem. In practice, efficient operators such as resizing or bit-depth reduction are used as empirical MAP realizations that approximate the anchoring effect by suppressing unstable off-manifold perturbation components while preserving the main semantic structure. 
Stronger generative operators can provide a closer manifold-anchoring effect at higher computational cost.
Different practical MAP instances correspond to different implicit or explicit choices of $\mathcal{R}_{\mathcal{M}}$, ranging from structural smoothing priors to learned denoising or generative purification priors. Integrating MAP yields the composite attack objective $\mathcal{L}(\mathcal{S}(\mathcal{P}_{\mathcal{M}}[\mathbf{x}+\boldsymbol{\delta}]), \mathbf{y})$, which regularizes the trajectory away from non-robust orthogonal directions and toward manifold-compatible adversarial vulnerabilities.

\noindent\textbf{Unified and Flexible Implementation.}
All practical MAP instances are incorporated through a unified operator interface,
$\tilde{\mathbf{x}}=\mathcal{P}_{\mathcal{M}}[\mathbf{x}+\boldsymbol{\delta}]$, before surrogate evaluation.
Thus, different implementations only instantiate $\mathcal{P}_{\mathcal{M}}$ in different ways, while the subsequent attack objective $\mathcal{L}(\mathcal{S}(\tilde{\mathbf{x}}),\mathbf{y})$ remains unchanged. 
This interface covers three categories:
\textbf{(i)} \textit{Explicit Priors}, such as spatial rescaling~\cite{guo2017countering}, bit-depth reduction, and frequency filtering, which provide low-cost empirical anchoring;
\textbf{(ii)} \textit{Discriminative Priors}, such as DnCNN~\cite{zhang2017beyond}, which implement anchoring through learned denoising; and
\textbf{(iii)} \textit{Generative Priors}, such as Latent Diffusion Models~\cite{rombach2022high}, which provide stronger purification-based anchoring at higher cost.
This unified view enables a flexible trade-off between anchoring strength and computational efficiency within the same manifold-anchored attack objective. More details are provided in \textit{\textbf{Appendix}~\ref{Method_Details}}.

\subsection{Distributional Bilevel Optimization}
\label{sec:mcba_framework}

Integrating the geometric constraints of $\mathcal{P}_{\mathcal{M}}$, we formulate transfer attack as DBO. 
By treating the adversarial initialization $\boldsymbol{\xi}$ as a learnable latent variable, we seek a geometry-aligned state $\boldsymbol{\xi}^{*}$ whose induced perturbation $\boldsymbol{\delta}^{*}(\boldsymbol{\xi})$ minimizes the expected transfer risk under surrogate uncertainty:
\begin{equation}
	\label{eq:bilevel_general}
	\min_{\boldsymbol{\xi} \in \Omega}~
	F(\boldsymbol{\xi}, \boldsymbol{\delta}^*(\boldsymbol{\xi})) := -\mathcal{R}_{dist}(\boldsymbol{\delta}^*(\boldsymbol{\xi}); \pi),~
	\text{s.t.}~ \boldsymbol{\delta}^*(\boldsymbol{\xi}) = \operatorname*{arg\,min}_{\boldsymbol{\delta} \in \Omega}~
	\underbrace{\left( - \mathcal{L} \left( \mathcal{S}\big(\mathcal{P}_{\mathcal{M}}[\mathbf{x} + \boldsymbol{\delta}]\big), \mathbf{y} \right) \right)}_{f(\boldsymbol{\delta}, \boldsymbol{\xi}; \mathcal{S})}.
\end{equation}
Here, $F(\cdot)$ and $f(\cdot)$ denote the upper-level and lower-level objectives, respectively. 
The mapping $\boldsymbol{\delta}^{*}(\boldsymbol{\xi})$ is the inner optimal perturbation obtained by solving $f(\cdot)$ from initialization $\boldsymbol{\delta}_{0}=\boldsymbol{\xi}$. 
Although $f$ is written as $f(\boldsymbol{\delta},\boldsymbol{\xi};\mathcal{S})$, its dependence on $\boldsymbol{\xi}$ arises through this initialization and the resulting manifold-anchored optimization trajectory.

\noindent\textbf{Lower-level Perturbation Generation.}
The lower-level problem optimizes $\boldsymbol{\delta}$ under the $\ell_p$ budget while evaluating the loss on the anchored state $\mathcal{P}_{\mathcal{M}}[\mathbf{x}+\boldsymbol{\delta}]$. 
This anchoring suppresses surrogate-specific off-manifold artifacts and biases the trajectory toward vulnerabilities that are more likely to remain shared across unknown architectures.

\noindent\textbf{Upper-level Risk Minimization.}
The upper-level objective $\mathcal{R}_{dist}$ bridges the white-box surrogate and unknown victims by optimizing against surrogate uncertainty. 
Inspired by the Bayesian perspective on transferability~\cite{li2023making}, we characterize a local hypothesis family around $\mathcal{S}$ via a surrogate-induced distribution $\pi(\boldsymbol{\theta}|\mathcal{S})$:
\begin{equation}
	\label{eq:rdist_def}
	\mathcal{R}_{dist}(\boldsymbol{\delta}; \pi) :=
	\mathbb{E}_{\boldsymbol{\theta} \sim \pi}
	\left[
	\mathcal{L}\left(
			\mathcal{S}_{\boldsymbol{\theta}}(\mathbf{x}+\boldsymbol{\delta}), \mathbf{y}
	\right)
	\right].
\end{equation}
In practice, $\pi(\boldsymbol{\theta}|\mathcal{S})$ is instantiated as a local Gaussian approximation around a fine-tuned surrogate checkpoint, rather than an exact distribution over unknown victims.
The expectation in Eq.~\eqref{eq:rdist_def} is estimated by averaging over a small set of sampled surrogate variants, with implementation details deferred to \textit{\textbf{Appendix}~\ref{Method_Details}}.
Minimizing this expectation prevents overfitting to a single deterministic boundary and guides $\boldsymbol{\xi}$ toward a distributionally transferable region where $\boldsymbol{\delta}^{*}(\boldsymbol{\xi})$ remains effective under surrogate variations.

\subsection{Hessian-Free Solution Strategy}

To avoid differentiating through the lower-level dynamics, we adopt a value-function approach based on the \emph{regularized gap function}~\cite{yaoovercoming}, which encodes lower-level optimality as a differentiable penalty.
Let $\mathcal{L}(\boldsymbol{\delta}) := \mathcal{L}(\mathcal{S}(\mathcal{P}_{\mathcal{M}}[\mathbf{x}+\boldsymbol{\delta}]), \mathbf{y})$. 
The gap function is defined as
\begin{equation}
	\label{eq:gap_function}
	\mathcal{G}_{\gamma}(\boldsymbol{\xi}, \boldsymbol{\delta})
	:= \max_{\hat{\boldsymbol{\delta}}\in\Omega}
	\left\{
	\mathcal{L}(\hat{\boldsymbol{\delta}})-\mathcal{L}(\boldsymbol{\delta})
	-\frac{1}{2\gamma}\|\hat{\boldsymbol{\delta}}-\boldsymbol{\delta}\|^2
	\right\}.
\end{equation}
When $\mathcal{G}_{\gamma}\!\to\!0$, $\boldsymbol{\delta}$ approaches a stationary point of the lower-level problem. 
We then optimize the gap-regularized meta objective as follows:
\begin{equation}
	\label{eq:joint_objective}
	\max_{\boldsymbol{\xi}\in\Omega}~
	\phi(\boldsymbol{\xi})
	:= \frac{1}{\lambda}\mathcal{R}_{dist}(\boldsymbol{\delta}^{*}(\boldsymbol{\xi});\pi)
	-\mathcal{G}_{\gamma}(\boldsymbol{\xi},\boldsymbol{\delta}^{*}(\boldsymbol{\xi})).
\end{equation}
Exact $\nabla_{\boldsymbol{\xi}}\phi$ requires backpropagation through the inner solver. 
We therefore use a first-order hyper-gradient approximation by identifying the dependence on $\boldsymbol{\xi}$ with the current inner state $\boldsymbol{\delta}^{*}$, i.e., $\nabla_{\boldsymbol{\xi}}\!\approx\!\nabla_{\boldsymbol{\delta}}$. 
By Danskin's theorem, a subgradient of $\mathcal{G}_{\gamma}$ is given by the maximizer $\hat{\boldsymbol{\delta}}^{*}$ in Eq.~\eqref{eq:gap_function}. 
We approximate it by one-step probing with radius $\sigma$,
$\hat{\boldsymbol{\delta}}=\Pi_{\Omega}(\boldsymbol{\delta}^{*}+\sigma\nabla_{\boldsymbol{\delta}}\mathcal{L}(\boldsymbol{\delta}^{*}))$, yielding
\begin{equation}
	\label{eq:hyper_gradient}
	\nabla_{\boldsymbol{\xi}}\phi
	\approx
	\frac{1}{\lambda}\nabla_{\boldsymbol{\delta}}\mathcal{R}_{dist}(\boldsymbol{\delta}^{*};\pi)
	-\left(\nabla_{\boldsymbol{\delta}}\mathcal{L}(\hat{\boldsymbol{\delta}})
	-\nabla_{\boldsymbol{\delta}}\mathcal{L}(\boldsymbol{\delta}^{*})\right).
\end{equation}
This estimator uses only gradients at $\boldsymbol{\delta}^{*}$ and $\hat{\boldsymbol{\delta}}$, together with $\nabla_{\boldsymbol{\delta}}\mathcal{R}_{dist}$, avoiding any Hessian-vector products. Alg.~\ref{alg:mabt} summarizes the resulting MABT procedure.

\noindent\textbf{Algorithm Analysis.}
We analyze solver stability using the projected gradient mapping
$\mathcal{G}_{proj}(\boldsymbol{\xi}_{t}) :=
\frac{1}{\alpha_t}(\Pi_{\Omega}(\boldsymbol{\xi}_{t}+\alpha_t\nabla\phi(\boldsymbol{\xi}_{t}))-\boldsymbol{\xi}_{t})$.
Detailed proofs are provided in \textit{\textbf{Appendix}~\ref{sec:proof_convergence}}.

\begin{theorem}[Non-asymptotic convergence]
	\label{thm:convergence}
	Under standard assumptions, let $\sigma=\Theta(1/\sqrt{T})$.
	With $\alpha_t=\Theta(1/\sqrt{T})$ and $\alpha_t\le\frac{1}{2L_{\phi}}$ for sufficiently large $T$, the sequence generated by MABT satisfies:
	\begin{equation}
		\min_{0\le t<T}\mathbb{E}\!\left[\|\mathcal{G}_{proj}(\boldsymbol{\xi}_t)\|^2\right]
		\le
		\mathcal{O}\!\left(\frac{1}{\sqrt{T}}\right)
		+\mathcal{O}\!\left(\frac{1}{T}\right)
		=\mathcal{O}\!\left(\frac{1}{\sqrt{T}}\right).
	\end{equation}
\end{theorem}


\noindent\textbf{Remark.}
Theorem~\ref{thm:convergence} does not claim global optimality of the non-convex bilevel problem; it guarantees projected first-order stationarity of the gap-regularized update under the smooth composite setting.
For practical MAP instances that are non-smooth or use surrogate gradients, the same update rule is applied with implementation details and empirical validation provided in the Appendix.
For clarity, the analysis uses the projection-based update and omits $\mathrm{sign}(\cdot)$, which is only used in implementation to match the standard $\ell_p$-budgeted attack routine; stationarity is defined by $\mathcal{G}_{proj}$.

\begin{algorithm2e}[t]
	\caption{Manifold Anchored Bilevel Transfer}
	\label{alg:mabt}
	\SetAlgoLined
	\DontPrintSemicolon
	\SetCommentSty{textnormal} 
	\KwIn{Image pair $(\mathbf{x}, \mathbf{y})$, surrogate $\mathcal{S}$, MAP $\mathcal{P}_{\mathcal{M}}$.
		Parameters: perturbation budget $\epsilon$; steps $T, K$; step sizes $\alpha, \beta$; probing radius $\sigma$; regularization coeff $\lambda$.}
	\KwOut{Adversarial example $\mathbf{x}_{adv}$.}
	
	\BlankLine
	\textbf{Initialize $\boldsymbol{\xi}_0 \gets \mathbf{0}$}\;
	
	\BlankLine
	\tcc{I: Manifold-Constrained Bilevel Learning}
	\For{$t = 0$ \KwTo $T-1$}{
		$\boldsymbol{\delta}_t \gets \boldsymbol{\xi}_t$\;
		
		$\mathbf{g}_t \gets \nabla_{\boldsymbol{\delta}} \mathcal{L}(\mathcal{S}(\mathcal{P}_{\mathcal{M}}[\mathbf{x} + \boldsymbol{\delta}_t]), \mathbf{y})$\tcp*[r]{Single-step manifold-anchored attack}
		
		$\boldsymbol{\delta}_{t+1} \gets \Pi_{\Omega}\!\left(\boldsymbol{\delta}_t + \beta \cdot \text{sign}(\mathbf{g}_t)\right)$\;
		
		$\mathbf{g}_{t+1} \gets \nabla_{\boldsymbol{\delta}} \mathcal{L}(\mathcal{S}(\mathcal{P}_{\mathcal{M}}[\mathbf{x} + \boldsymbol{\delta}_{t+1}]), \mathbf{y})$\tcp*[r]{Probing to approximate the gap maximizer}
		
		$\hat{\boldsymbol{\delta}} \gets \Pi_{\Omega}\!\left(\boldsymbol{\delta}_{t+1} + \sigma \cdot \mathbf{g}_{t+1}\right)$\;
		
		$\hat{\mathbf{g}} \gets \nabla_{\boldsymbol{\delta}} \mathcal{L}(\mathcal{S}(\mathcal{P}_{\mathcal{M}}[\mathbf{x} + \hat{\boldsymbol{\delta}}]), \mathbf{y})$\tcp*[r]{Gap gradient estimator}
		
		$\nabla \mathcal{G} \gets \hat{\mathbf{g}} - \mathbf{g}_{t+1}$\;
		
		$\mathbf{v} \gets \nabla_{\boldsymbol{\delta}} \mathcal{R}_{dist}(\boldsymbol{\delta}_{t+1}; \pi)$\tcp*[r]{Distributional transfer risk gradient}
		
		$\boldsymbol{\xi}_{t+1} \gets \Pi_{\Omega}\!\left(\boldsymbol{\xi}_t + \alpha \cdot \text{sign}\!\left(\frac{1}{\lambda}\mathbf{v} - \nabla \mathcal{G}\right)\right)$\tcp*[r]{Meta update: maximize $\Phi=\frac{1}{\lambda}\mathcal{R}_{dist}-\mathcal{G}_\gamma$}
	}
	
	\BlankLine
	\tcc{II: Manifold Adversarial Adaptation}
	\textbf{Reset:} $\boldsymbol{\delta}_0 \gets \boldsymbol{\xi}_T$\;
	\For{$k = 0$ \KwTo $K-1$}{
		
		$\boldsymbol{\delta}_{k+1} \gets 
		\Pi_{\Omega}\!\left(
		\boldsymbol{\delta}_k + \alpha \cdot 
		\text{sign}\!\left(
		\nabla_{\boldsymbol{\delta}}\mathcal{L}
		\big(\mathcal{S}(\mathcal{P}_{\mathcal{M}}[\mathbf{x}+\boldsymbol{\delta}_k]), \mathbf{y}\big)
		\right)
		\right)$\;
		
	}
	\KwRet $\mathbf{x}_{adv} = \mathbf{x} + \boldsymbol{\delta}_K$\;
\end{algorithm2e}

%
%

\begin{table*}[t]
	\centering
	\caption{ASR (\%) of MABT across  \textbf{16} attack configurations on the ResNet-50 surrogate. Results cover 4 base attackers combined with ensemble/model-related strategies. Best results are in \textbf{bold}.}
	\label{tab:main_results}
	\scalebox{0.97}{
		\begin{threeparttable}
			\fontsize{7.pt}{8.pt}\selectfont
			\renewcommand\arraystretch{1.1} 
			\setlength{\aboverulesep}{0.5pt}
			\setlength{\belowrulesep}{0.5pt}
			\setlength{\tabcolsep}{0.85pt}
			
			\begin{tabular}{ll cccccc ccc ccc c}
				\toprule[0.8pt]
				\multicolumn{15}{c}{\textbf{Untargeted Attack Scenario, ResNet-50 backbone, Average Success Rate (\%) $\uparrow$}} \\
				\cmidrule(lr){1-15} 
				
				\multicolumn{2}{c}{\multirow{2}{*}{\textbf{Method}}} & 
				\multicolumn{6}{c}{\textbf{CNN}} & 
				\multicolumn{3}{c}{\textbf{CNN Ensemble}} & 
				\multicolumn{3}{c}{\textbf{Transformer}} &
				\multirow{2}{*}{\textbf{Avg.}} \\
				
				\cmidrule(lr){3-8} \cmidrule(lr){9-11} \cmidrule(lr){12-14}
				
				& & Inc-v3 & Inc-Res-v2 & DenseNet & MobileNet & PNASNet & SENet & $\text{Inc-v3}_{\text{ens3}}$ & $\text{Inc-v3}_{\text{ens4}}$ & $\text{IncRes-v2}_{\text{ens}}$ & Visformer-s &DeiT-s & ConViT-b & \\
				\midrule
				
				& N/A & 16.38 & 13.04 & 37.52 & 36.50 & 12.80 & 17.10 & 10.18 & 9.40 & 5.86 & 10.98 & 7.50 & 5.56 & 15.24 \\
				\rowcolor{ourscore}
				\cellcolor{white} & \textbf{Ours} & \textbf{34.96} & \textbf{24.58} & \textbf{70.00} & \textbf{69.22} & \textbf{27.84} & \textbf{33.94} & \textbf{19.72} & \textbf{16.38} & \textbf{10.28} & \textbf{17.02} & \textbf{11.54} & \textbf{7.62} & \textbf{28.59} \\
				
				& GHOST & 18.34 & 14.44 & 41.56 & 41.72 & 13.60 & 19.78 & 10.98 & 10.00 & 6.02 & 12.10 & 7.14 & 5.90 & 16.80 \\
				\rowcolor{ourscore}
				\cellcolor{white} & \textbf{Ours} & \textbf{37.38} & \textbf{27.52} & \textbf{74.66} & \textbf{72.58} & \textbf{30.84} & \textbf{36.72} & \textbf{21.18} & \textbf{17.66} & \textbf{10.50} & \textbf{18.24} & \textbf{11.92} & \textbf{7.72} & \textbf{30.58} \\
				
				& MBA & 21.66 & 14.98 & 49.22 & 57.74 & 13.06 & 20.22 & 12.28 & 10.26 & 5.94 & 11.00 & 7.58 & 4.52 & 19.04 \\
				\rowcolor{ourscore}
				\cellcolor{white} & \textbf{Ours} & \textbf{44.20} & \textbf{32.52} & \textbf{78.76} & \textbf{81.22} & \textbf{32.60} & \textbf{39.20} & \textbf{27.12} & \textbf{22.44} & \textbf{14.54} & \textbf{19.08} & \textbf{13.80} & \textbf{8.36} & \textbf{34.49} \\
				
				& AWT & 21.24 & 15.92 & 42.30 & 42.38 & 15.58 & 18.50 & 13.52 & 13.10 & 8.10 & 12.04 & 8.78 & 6.32 & 18.15 \\
				\rowcolor{ourscore}
				\cellcolor{white}\multirow{-8}{*}{\textbf{PGD}} & \textbf{Ours} & \textbf{33.94} & \textbf{24.58} & \textbf{59.56} & \textbf{61.92} & \textbf{24.68} & \textbf{28.60} & \textbf{22.52} & \textbf{20.76} & \textbf{13.52} & \textbf{15.52} & \textbf{12.58} & \textbf{8.12} & \textbf{27.19} \\
				\midrule
				
				& N/A & 26.58 & 21.76 & 51.36 & 49.54 & 22.96 & 30.26 & 15.92 & 14.54 & 8.98 & 17.96 & 11.96 & 9.04 & 23.41 \\
				\rowcolor{ourscore}
				\cellcolor{white} & \textbf{Ours} & \textbf{59.62} & \textbf{47.92} & \textbf{88.78} & \textbf{87.48} & \textbf{49.92} & \textbf{58.54} & \textbf{37.74} & \textbf{32.08} & \textbf{21.02} & \textbf{34.86} & \textbf{22.44} & \textbf{15.92} & \textbf{46.36} \\
				
				& GHOST & 30.10 & 24.02 & 58.28 & 55.82 & 24.86 & 34.00 & 17.00 & 15.12 & 9.86 & 19.40 & 12.34 & 9.06 & 25.82 \\
				\rowcolor{ourscore}
				\cellcolor{white} & \textbf{Ours} & \textbf{63.68} & \textbf{51.10} & \textbf{91.22} & \textbf{90.44} & \textbf{53.88} & \textbf{62.42} & \textbf{40.14} & \textbf{33.64} & \textbf{22.72} & \textbf{37.20} & \textbf{24.40} & \textbf{16.40} & \textbf{48.94} \\
				
				& MBA & 40.42 & 29.78 & 72.62 & 78.42 & 27.72 & 37.60 & 23.32 & 19.90 & 12.10 & 21.68 & 14.04 & 9.26 & 32.24 \\
				\rowcolor{ourscore}
				\cellcolor{white} & \textbf{Ours} & \textbf{67.28} & \textbf{55.56} & \textbf{93.00} & \textbf{92.86} & \textbf{56.52} & \textbf{63.10} & \textbf{46.84} & \textbf{40.76} & \textbf{28.42} & \textbf{40.22} & \textbf{27.42} & \textbf{18.70} & \textbf{52.56} \\
				
				& AWT & 32.42 & 25.82 & 56.02 & 55.40 & 27.16 & 30.36 & 21.56 & 19.74 & 13.94 & 18.02 & 13.76 & 9.82 & 27.00 \\
				\rowcolor{ourscore}
				\cellcolor{white} \multirow{-8}{*}{\textbf{MI}} & \textbf{Ours} & \textbf{54.24} & \textbf{43.10} & \textbf{81.44} & \textbf{82.00} & \textbf{43.54} & \textbf{48.90} & \textbf{37.48} & \textbf{33.70} & \textbf{23.78} & \textbf{30.18} & \textbf{22.12} & \textbf{15.02} & \textbf{42.96} \\
				\midrule
				
				& N/A & 17.30 & 14.48 & 40.10 & 38.18 & 15.98 & 19.80 & 11.24 & 10.80 & 6.48 & 11.50 & 8.50 & 6.32 & 16.72 \\
				\rowcolor{ourscore}
				\cellcolor{white} & \textbf{Ours} & \textbf{36.22} & \textbf{25.38} & \textbf{70.08} & \textbf{68.32} & \textbf{29.58} & \textbf{33.10} & \textbf{22.12} & \textbf{19.00} & \textbf{12.14} & \textbf{15.36} & \textbf{11.86} & \textbf{7.24} & \textbf{29.20} \\
				
				& GHOST & 19.92 & 15.58 & 44.76 & 43.26 & 17.30 & 22.52 & 12.00 & 10.76 & 6.70 & 12.38 & 8.68 & 6.50 & 18.36 \\
				\rowcolor{ourscore}
				\cellcolor{white} & \textbf{Ours} & \textbf{36.56} & \textbf{25.78} & \textbf{72.32} & \textbf{70.04} & \textbf{30.16} & \textbf{34.80} & \textbf{22.22} & \textbf{19.26} & \textbf{12.22} & \textbf{16.02} & \textbf{11.94} & \textbf{7.24} & \textbf{29.88} \\
				
				& MBA & 24.66 & 18.22 & 55.76 & 60.20 & 17.46 & 23.72 & 14.78 & 13.20 & 8.60 & 11.80 & 9.24 & 5.62 & 21.94 \\
				\rowcolor{ourscore}
				\cellcolor{white} & \textbf{Ours} & \textbf{44.82} & \textbf{32.28} & \textbf{78.48} & \textbf{79.34} & \textbf{34.76} & \textbf{38.40} & \textbf{29.52} & \textbf{25.94} & \textbf{17.24} & \textbf{17.80} & \textbf{14.82} & \textbf{8.50} & \textbf{35.16} \\
				
				& AWT & 24.22 & 18.46 & 47.44 & 46.50 & 19.58 & 21.64 & 16.16 & 15.38 & 10.34 & 12.80 & 10.22 & 7.20 & 20.83 \\
				\rowcolor{ourscore}
				\cellcolor{white} \multirow{-8}{*}{\textbf{TI}} & \textbf{Ours} & \textbf{36.18} & \textbf{25.56} & \textbf{62.04} & \textbf{63.88} & \textbf{27.58} & \textbf{29.42} & \textbf{25.32} & \textbf{23.88} & \textbf{16.56} & \textbf{15.38} & \textbf{13.14} & \textbf{8.54} & \textbf{28.96} \\
				\midrule
				
				& N/A & 31.46 & 24.98 & 57.76 & 56.68 & 24.42 & 27.56 & 21.76 & 20.02 & 12.80 & 18.60 & 13.54 & 9.86 & 26.62 \\
				\rowcolor{ourscore}
				\cellcolor{white} & \textbf{Ours} & \textbf{59.30} & \textbf{50.86} & \textbf{85.72} & \textbf{82.34} & \textbf{58.12} & \textbf{58.14} & \textbf{43.64} & \textbf{39.00} & \textbf{29.44} & \textbf{39.60} & \textbf{26.48} & \textbf{19.34} & \textbf{49.33} \\
				
				& GHOST & 35.20 & 27.04 & 64.24 & 63.72 & 27.60 & 31.14 & 24.64 & 22.48 & 14.80 & 21.24 & 14.56 & 10.94 & 29.80 \\
				\rowcolor{ourscore}
				\cellcolor{white} & \textbf{Ours} & \textbf{67.16} & \textbf{59.24} & \textbf{91.30} & \textbf{88.76} & \textbf{65.78} & \textbf{66.78} & \textbf{50.98} & \textbf{45.44} & \textbf{33.86} & \textbf{47.72} & \textbf{30.72} & \textbf{22.72} & \textbf{55.87} \\
				
				& MBA & 33.74 & 23.44 & 65.86 & 73.78 & 22.02 & 29.68 & 21.00 & 19.28 & 11.62 & 17.76 & 11.30 & 7.10 & 28.05 \\
				\rowcolor{ourscore}
				\cellcolor{white} & \textbf{Ours} & \textbf{63.84} & \textbf{51.42} & \textbf{90.12} & \textbf{90.24} & \textbf{53.68} & \textbf{57.10} & \textbf{47.42} & \textbf{42.52} & \textbf{31.18} & \textbf{34.26} & \textbf{24.96} & \textbf{16.58} & \textbf{50.28} \\
				
				& AWT & 24.80 & 17.46 & 46.90 & 46.76 & 16.88 & 18.16 & 17.06 & 16.52 & 10.22 & 11.94 & 10.30 & 7.12 & 20.34 \\
				\rowcolor{ourscore}
				\cellcolor{white} \multirow{-8}{*}{\textbf{GAA}} & \textbf{Ours} & \textbf{47.78} & \textbf{37.98} & \textbf{71.90} & \textbf{70.14} & \textbf{39.52} & \textbf{39.78} & \textbf{35.50} & \textbf{34.28} & \textbf{25.02} & \textbf{23.54} & \textbf{19.56} & \textbf{13.02} & \textbf{38.17} \\
				
				\bottomrule[0.8pt]
			\end{tabular}
		\end{threeparttable}
	}
	    \vspace{-0.4cm}
\end{table*}

\section{Experiments}

\textbf{Dataset and Models.} 
We follow commonly used settings to select 5,000 correctly classified images from the ImageNet validation set. 
To evaluate cross-architecture generalization, we employ ResNet-50 and VGG-19 as surrogates. 
The generated AEs are evaluated on 12 diverse victim models spanning CNNs, robust ensembles, and Transformers: 
\textit{(i)} \textbf{6 Standard CNNs}: Inception-v3 (Inc-v3)~\cite{szegedy2016rethinking}, IncRes-v2~\cite{szegedy2017inception}, DenseNet~\cite{huang2017densely}, MobileNet~\cite{sandler2018mobilenetv2}, PNASNet~\cite{liu2018progressive}, and SENet~\cite{hu2018squeeze}; 
\textit{(ii)} \textbf{3 Robust Ensembles}: Inc-v3$_{ens3}$, Inc-v3$_{ens4}$, and IncRes$_{ens}$~\cite{tramer2017ensemble}; 
and \textit{(iii)} \textbf{3 Transformers}: Visformer~\cite{chen2021visformer}, DeiT-S~\cite{touvron2021training}, and ConViT-B~\cite{d2021convit}. 
We use Attack Success Rate (ASR) as the primary metric.

\textbf{Baselines and Implementation Details.} 
We evaluate MABT as a flexible framework integrated with diverse attackers. 
We use gradient-based methods (PGD~\cite{kurakin2018adversarial}, MI~\cite{dong2018boosting}, VMI~\cite{wang2021enhancing}, GAA~\cite{gan2025boosting}) and input-transformation methods (SI~\cite{lin2019nesterov}, DI~\cite{xie2019improving}, TI~\cite{dong2019evading}) as \textit{base attackers}.  
We further combine them with ensemble-based methods (GHOST~\cite{li2020learning}, MBA~\cite{li2023making}) and model-related strategies (AWT~\cite{chen2025enhancing}), covering the major design paradigms of transfer attacks. 
We also compare BETAK~\cite{liu2024advancing} as a recent initialization-based transfer attack. 
Finally, we evaluate transferability against five defenses: HGD~\cite{liao2018defense}, R\&P~\cite{xie2017mitigating}, NIPS-r3~\cite{kurakin2018adversarial2}, JPEG~\cite{guo2017countering}, and RS~\cite{cohen2019certified}.  

\textit{Hyperparameters:} 
We adopt BETAK's official settings with Inc-v3 as its white-box pseudo-victim. 
For all methods, we set $\epsilon=0.03$, $K=10$, and $\alpha=0.006$.  
We use spatial resizing (Resize) as the default MAP instantiation, as it provides the best efficiency--performance trade-off according to the analysis in Sec.~4.2. 
For MABT, we set $\sigma=0.01$ following~\cite{yaoovercoming}.  
The iteration $T=3$, regularization $\lambda=0.01$, and lower-level step size $\beta=1.0$ are selected based on ablation results.

\subsection{Experimental Results}

\begin{wrapfigure}{r}{7cm}
	\vspace{-0.5cm} %
	\begin{minipage}[t]{1.0\linewidth}
		\centering
		\makeatletter\def\@captype{table}\makeatother\caption{Comparison with initialization-based BETAK which uses auxiliary \textit{pseudo-victims} (Inc-v3).}
		\label{tab:betak_results}
		\scalebox{0.95}{
			\begin{threeparttable}
				\fontsize{7.5pt}{8.5pt}\selectfont
				\renewcommand\arraystretch{1.} 
				\setlength{\aboverulesep}{0.5pt}
				\setlength{\belowrulesep}{0.5pt}
				\setlength{\tabcolsep}{3pt} 
				\begin{tabular}{ll cccc}
					\toprule[0.8pt]
					\multicolumn{2}{c}{\multirow{2}{*}{\textbf{Method}}} & 
					\multicolumn{3}{c}{\textbf{Average ASR (\%)$\uparrow$, ResNet-50}} & 
					\multirow{2}{*}{\textbf{Avg.}} \\
					\cmidrule(lr){3-5}
					& & \textbf{CNN} & \textbf{CNN Ensemble} & \textbf{Transformer} & \\
					\midrule
					
					\multirow{3}{*}{\textbf{PGD}} & N/A & 22.22 & 8.48 & 8.01 & 15.24 \\
					& BETAK & 30.87 & 11.75 & 10.47 & 20.99 \\
					\rowcolor{ourscore}
					\cellcolor{white} & \textbf{Ours} & \textbf{43.42} & \textbf{15.46} & \textbf{12.06} & \textbf{28.59} \\
					\cmidrule(lr){2-6}
					
					\multirow{3}{*}{\textbf{MI}} & N/A & 33.74 & 13.15 & 12.99 & 23.41 \\
					& BETAK & 43.86 & 19.74 & 15.97 & 30.86 \\
					\rowcolor{ourscore}
					\cellcolor{white} & \textbf{Ours} & \textbf{65.38} & \textbf{30.28} & \textbf{24.41} & \textbf{46.36} \\
					\cmidrule(lr){2-6}
					
					\multirow{3}{*}{\textbf{TI}} & N/A & 24.31 & 9.51 & 8.77 & 16.72 \\
					& BETAK & 32.41 & 13.12 & 10.75 & 22.17 \\
					\rowcolor{ourscore}
					\cellcolor{white} & \textbf{Ours} & \textbf{43.78} & \textbf{17.75} & \textbf{11.49} & \textbf{29.20} \\
					\cmidrule(lr){2-6}
					
					\multirow{3}{*}{\textbf{GAA}} & N/A & 37.14 & 18.19 & 14.00 & 26.62 \\
					& BETAK & 45.55 & 23.49 & 18.46 & 33.26 \\
					\rowcolor{ourscore}
					\cellcolor{white} & \textbf{Ours} & \textbf{65.75} & \textbf{37.36} & \textbf{28.47} & \textbf{49.33} \\
					
					\bottomrule[0.8pt]
				\end{tabular}
			\end{threeparttable}
		}
	\end{minipage}
	\vspace{-0.8cm} %
\end{wrapfigure}

We provide additional results in the \textit{\textbf{Appendix}}, including \textbf{MAP and DBO ablations} (Tabs.~\ref{tab:ablation_study_MAP} and~\ref{tab:ablation_study_full}), \textbf{hyperparameter sensitivity} (Fig.~\ref{fig:hyperparameter}), \textbf{additional attackers and VGG-19} (Tabs.~\ref{tab:main_results_full} and~\ref{tab:vgg_results}), and \textbf{targeted attacks} (Tab.~\ref{tab:target_results}).

\noindent\textbf{Quantitative Comparison.}
As shown in Tab.~\ref{tab:main_results}, MABT raises the overall average ASR from {22.52\%} to {39.28\%}, corresponding to a \textbf{74.4\%} relative gain, with particularly clear gains on heterogeneous ViTs and robust ensembles. Results on the VGG-19 surrogate (Tab.~\ref{tab:vgg_results} in \textit{\textbf{Appendix}}) further support the effectiveness of MABT under a structurally distinct surrogate. Together, these results demonstrate that MABT can serve as a unified transfer framework across different surrogate choices and victim architectures.
\begin{table*}[htbp]
	\centering
	\caption{ASR (\%) results against five representative defenses using the ResNet-50 surrogate.}
	\scalebox{0.97}{
		\begin{threeparttable}
			\fontsize{7.5pt}{8.5pt}\selectfont
			\renewcommand\arraystretch{1.1} 
			\setlength{\aboverulesep}{0.5pt}
			\setlength{\belowrulesep}{0.5pt}
			\setlength{\tabcolsep}{2.85pt}
			
			\begin{tabular}{clcccccc c clcccccc}
				\toprule[0.8pt]
				\multicolumn{17}{c}{\textbf{Defense Mechanism, ResNet-50 backbone, Average Success Rate (\%)$\uparrow$}} \\
				\cmidrule(lr){1-17} 
				\multicolumn{2}{c}{\textbf{Method}}& HGD & JPEG & RS & R\&P & NIPS-r3 & \textbf{Avg.} & & \multicolumn{2}{c}{\textbf{Method}}& HGD & JPEG & RS & R\&P & NIPS-r3 & \textbf{Avg.}\\ 
				\midrule
				
				& N/A & 16.94 & 11.60 & 7.96 & 12.12 & 14.28 & 12.58 & & & N/A & 27.12 & 17.54 & 9.44 & 13.40 & 23.10 & 18.12 \\
				
				\rowcolor{ourscore}
				\cellcolor{white} & \textbf{Ours} & \textbf{36.16} & \textbf{21.38} & \textbf{10.42} & \textbf{13.90} & \textbf{29.32} & \textbf{22.24} & \cellcolor{white} & \cellcolor{white} & \textbf{Ours} & \textbf{59.82} & \textbf{39.68} & \textbf{14.80} & \textbf{18.28} & \textbf{53.68} & \textbf{37.25} \\

				& GHOST & 19.34 & 11.84 & 8.24 & 12.70 & 15.76 & 13.58 & & & GHOST & 30.94 & 19.02 & 9.40 & 13.92 & 24.80 & 19.62 \\
				
				\rowcolor{ourscore}
				\cellcolor{white} & \textbf{Ours} & \textbf{38.86} & \textbf{21.94} & \textbf{10.40} & \textbf{14.26} & \textbf{31.44} & \textbf{23.38} & \cellcolor{white} & \cellcolor{white} & \textbf{Ours} & \textbf{63.02} & \textbf{41.56} & \textbf{15.08} & \textbf{19.40} & \textbf{56.08} & \textbf{39.03} \\
				
				& MBA & 22.22 & 13.22 & 9.22 & 12.84 & 19.80 & 15.46 & & & MBA & 40.10 & 25.82 & 11.94 & 15.52 & 35.80 & 25.84 \\
				
				\rowcolor{ourscore}
				\cellcolor{white} & \textbf{Ours} & \textbf{43.98} & \textbf{28.48} & \textbf{11.08} & \textbf{14.84} & \textbf{40.10} & \textbf{27.70} & \cellcolor{white} & \cellcolor{white} & \textbf{Ours} & \textbf{67.04} & \textbf{48.80} & \textbf{16.50} & \textbf{19.18} & \textbf{63.50} & \textbf{43.00} \\
				
				& AWT & 20.94 & 16.28 & 10.30 & 14.54 & 19.90 & 16.39 & & & AWT & 32.58 & 25.60 & 13.18 & 17.46 & 30.18 & 23.80 \\
				
				\rowcolor{ourscore}
				\cellcolor{white} \multirow{-8}{*}{\textbf{PGD}}& \textbf{Ours} & \textbf{33.80} & \textbf{25.70} & \textbf{13.28} & \textbf{17.06} & \textbf{30.68} & \textbf{24.10} & \cellcolor{white} & \cellcolor{white} \multirow{-8}{*}{\textbf{MI}}& \textbf{Ours} & \textbf{54.82} & \textbf{41.88} & \textbf{18.20} & \textbf{22.58} & \textbf{51.32} & \textbf{37.76} \\
				
				\midrule
				
				& N/A & 19.62 & 12.70 & 8.44 & 11.88 & 15.86 & 13.70 & & & N/A & 32.10 & 24.34 & 13.10 & 18.02 & 29.14 & 23.34 \\
				
				\rowcolor{ourscore}
				\cellcolor{white} & \textbf{Ours} & \textbf{37.42} & \textbf{23.92} & \textbf{11.56} & \textbf{14.82} & \textbf{30.42} & \textbf{23.63} & \cellcolor{white} & \cellcolor{white} & \textbf{Ours} & \textbf{60.34} & \textbf{43.60} & \textbf{15.00} & \textbf{20.40} & \textbf{55.26} & \textbf{38.92} \\

				& GHOST & 21.22 & 13.26 & 8.66 & 11.86 & 17.76 & 14.55 & & & GHOST & 36.80 & 27.20 & 14.14 & 19.64 & 32.78 & 26.11 \\
				
				\rowcolor{ourscore}
				\cellcolor{white} & \textbf{Ours} & \textbf{39.20} & \textbf{23.98} & \textbf{11.76} & \textbf{14.92} & \textbf{31.22} & \textbf{24.22} & \cellcolor{white} & \cellcolor{white} & \textbf{Ours} & \textbf{68.12} & \textbf{50.70} & \textbf{15.70} & \textbf{21.54} & \textbf{62.86} & \textbf{43.78} \\
				
				& MBA & 27.06 & 16.18 & 10.24 & 13.16 & 23.76 & 18.08 & & & MBA & 32.48 & 23.24 & 12.58 & 17.20 & 31.04 & 23.31 \\
				
				\rowcolor{ourscore}
				\cellcolor{white}& \textbf{Ours} & \textbf{45.68} & \textbf{31.44} & \textbf{12.32} & \textbf{16.18} & \textbf{40.16} & \textbf{29.16} & \cellcolor{white} & \cellcolor{white} & \textbf{Ours} & \textbf{63.60} & \textbf{47.90} & \textbf{15.90} & \textbf{19.56} & \textbf{61.10} & \textbf{41.61} \\
				
				& AWT & 25.44 & 19.32 & 11.80 & 15.48 & 23.26 & 19.06 & & & AWT & 38.20 & 30.28 & 15.12 & 20.86 & 35.24 & 27.94 \\
				
				\rowcolor{ourscore}
				\cellcolor{white} \multirow{-8}{*}{\textbf{TI}} & \textbf{Ours} & \textbf{37.10} & \textbf{29.46} & \textbf{14.60} & \textbf{18.52} & \textbf{33.92} & \textbf{26.72} & \cellcolor{white} & \cellcolor{white} \multirow{-8}{*}{\textbf{GAA}}& \textbf{Ours} & \textbf{47.60} & \textbf{40.22} & \textbf{19.58} & \textbf{24.74} & \textbf{44.40} & \textbf{35.31} \\
				
				\bottomrule[0.8pt]
			\end{tabular}
		\end{threeparttable}
	}
	
	\label{tab:defense_combined_style}
\end{table*}

\begin{figure*}[!t]
	\centering
	\includegraphics[width=0.99\textwidth]{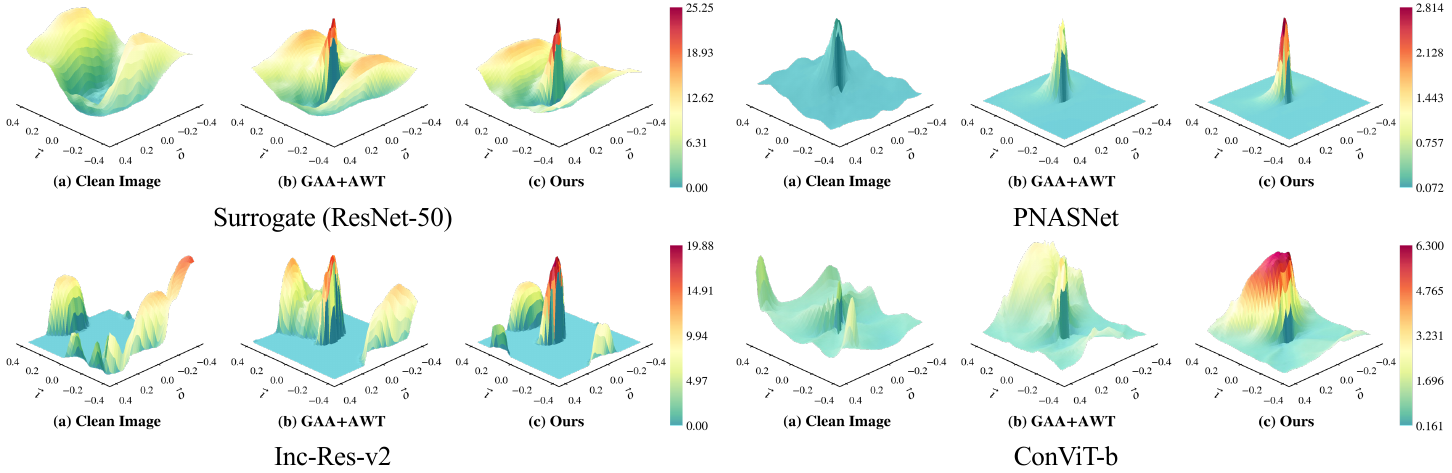} 
	\caption{
		\textbf{Adversarial loss landscape}. Surfaces are plotted along the \textit{gradient direction} and a \textit{random orthogonal direction}. 
		Unlike baselines that suffer from landscape collapse on unknown victims, MABT preserves higher victim-side loss basins with relatively flatter surrounding curvature across architectures, which indicates that the learned trajectory better retains transfer-relevant directions.}
	\label{fig:landscape}
	\vspace{-0.5cm}
\end{figure*}

\noindent\textbf{Visualization of Adversarial Loss Landscapes.} 
To explicate the mechanism of MABT, Fig.~\ref{fig:landscape} visualizes the adversarial loss landscape $\mathcal{L}(\mathbf{x} + a\boldsymbol{\vec{\iota}}+b\boldsymbol{\vec{o}})$, plotted along the gradient direction ${\vec{\iota}}$ and a random orthogonal direction ${\vec{o}}$. 
While mainstream attacks (e.g., GAA+AWT) generate sharp peaks on the surrogate that collapse rapidly on victim models (indicating severe overfitting), MABT maintains a higher loss basin with flatter curvature across architectures. This visualization supports our trajectory-alignment interpretation: MABT tends to preserve higher victim-side loss basins with flatter surrounding curvature, suggesting reduced surrogate-specific overfitting.

\noindent\textbf{Comparison with Initialization Attackers.} 
We also benchmark MABT against BETAK~\cite{liu2024advancing}. 
While BETAK relies on an auxiliary \textit{pseudo-victim} (i.e., Inc-v3), MABT operates without extra victim model dependencies, anchoring the adversarial trajectory through the intrinsic data geometry. To maintain a consistent evaluation protocol, the mean ASR of the CNN victim suite incorporates results from Inc-v3. On GAA, Tab.~\ref{tab:betak_results} shows that MABT increases the average ASR from 33.26\% for BETAK to 49.33\%, corresponding to a 48.32\% relative improvement and supporting the effectiveness of intrinsic geometric alignment.

\textbf{Evaluation against Defense Mechanisms.} We subsequently assess MABT under five representative defense strategies. As reported in Tab.~\ref{tab:defense_combined_style}, MABT improves the aggregate ASR under these defenses. These results suggest that manifold-anchored perturbations remain more transferable under diverse defense mechanisms.



\noindent\textbf{Targeted Attack Performance.} 
We further evaluate MABT on the more challenging targeted attack scenario. 
As shown in Tab.~\ref{tab:target_results} in \textit{\textbf{Appendix}}, MABT consistently outperforms baselines across diverse settings. This suggests that the geometry-aware optimization can also benefit targeted transfer by guiding perturbations toward target-relevant feature directions.

\subsection{Ablation Results}\label{sec:ablation_results_main}

\begin{figure*}[t]
	\centering
	\includegraphics[width=0.99\linewidth]{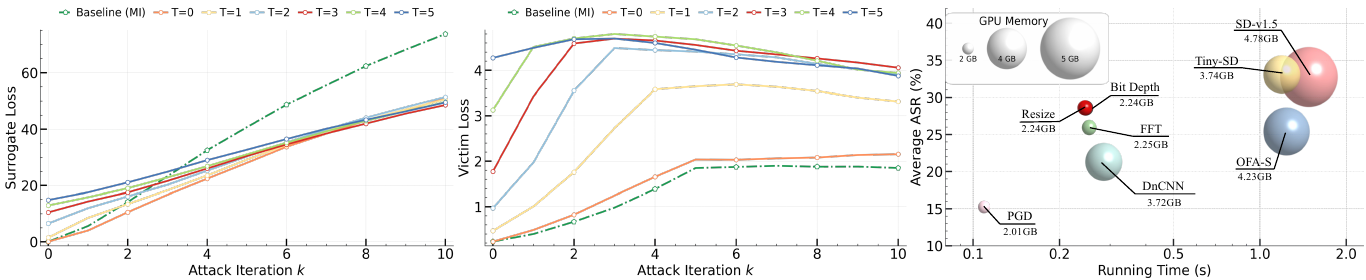} 
	\caption{
		\textbf{Convergence and efficiency analysis.} 
		\textbf{Left \& Mid:} Increasing $T$ improves geometry-aligned initialization to mitigate surrogate overfitting, thereby ensuring sustained victim loss growth during the subsequent attack.
		\textbf{Right:} Resize provides a favorable low-cost MAP realization, while generative priors achieve stronger anchoring at higher computational cost.
	}
	\label{fig:loss-figure}
	\vspace{-0.5cm}
\end{figure*}

\textbf{Convergence Analysis.}
Fig.~\ref{fig:loss-figure} shows distinct optimization behaviors.
PGD rapidly improves the surrogate loss but saturates early on the Inc-v3 victim, indicating severe surrogate-specific overfitting.
In contrast, larger $T$ in MABT keeps the surrogate trajectory controlled while increasing the victim-side loss, suggesting better transfer-oriented initialization.
Even $T=0$ (MAP-only) outperforms PGD, and $T=3$ provides a strong trade-off between warm-up strength and transferability.

\begin{wrapfigure}{r}{7cm}
	\vspace{-0.4cm} %
	\begin{minipage}[t]{1.0\linewidth}
		\centering
		\includegraphics[width=0.99\linewidth]{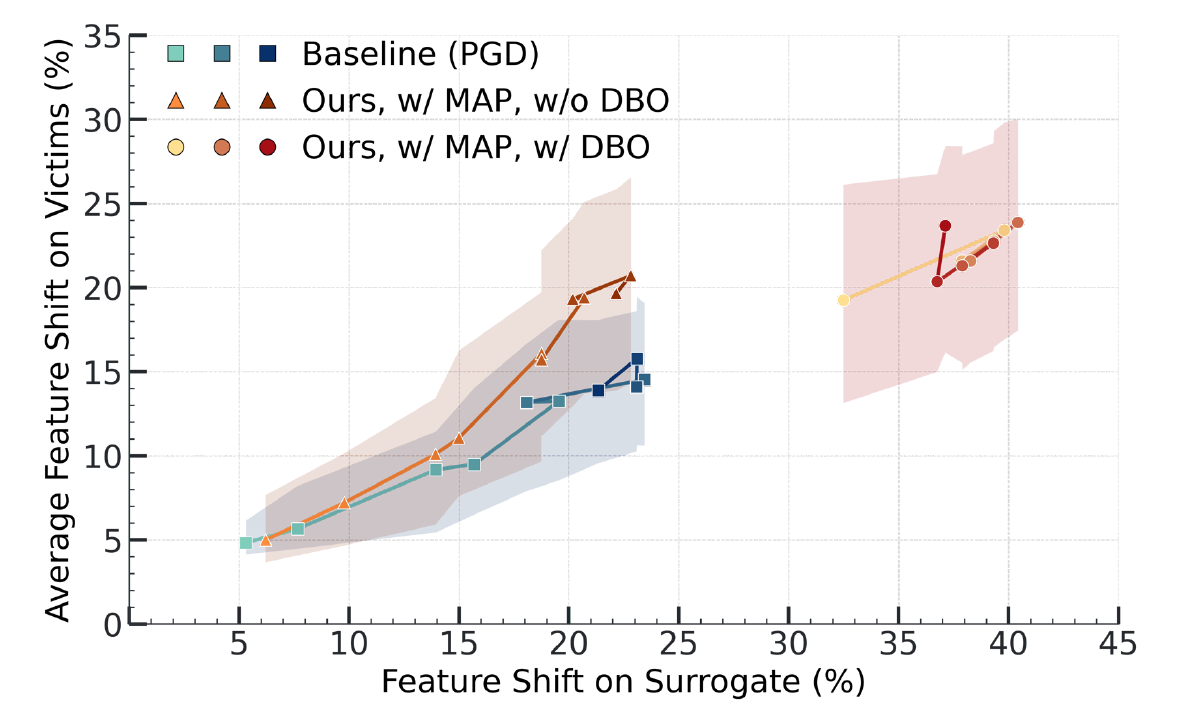} 
		\caption{\textbf{Feature shift consistency.} We plot surrogate-side (ResNet-50) feature shift against the average victim-side shift. Shaded regions denote the standard deviation over four heterogeneous victims (Inc-v3, DenseNet, DeiT-s and ConViT-b).}
		\label{fig:shadow}
	\end{minipage}
	\vspace{-0.3cm} %
\end{wrapfigure}

\textbf{Effectiveness of MAP and DBO.}
We evaluate each component through quantitative ablation and geometric feature-shift correlation.
\textit{(i)} \textbf{Component Gain.}
As shown in Tab.~\ref{tab:ablation_aggregated}, MAP and DBO each improve the baseline, and their combination achieves the best performance.
\textit{(ii)} \textbf{Feature-space Alignment.}
Fig.~\ref{fig:shadow} compares surrogate-side and victim-side feature shifts across heterogeneous victims.
PGD shows weaker correlation and larger variance, suggesting that stronger surrogate distortion does not reliably transfer to victims.
MAP improves this correlation by rectifying the trajectory, while DBO further shifts the initialization toward more transferable regions.
Together, the two components yield more consistent surrogate-victim feature-shift alignment across architectures, indicating that MABT better preserves transfer-relevant trajectory directions.

\begin{wrapfigure}{r}{7cm}
		\vspace{-0.4cm} %
	\begin{minipage}[t]{1.0\linewidth}
		\centering
		\makeatletter\def\@captype{table}\makeatother\caption{Ablation results by implementing DBO and MAP based on  PGD and PGD+AWT.}
		\label{tab:ablation_aggregated}
		
		\scalebox{0.97}{
			\begin{threeparttable}
				\fontsize{7.5pt}{8.5pt}\selectfont
				\renewcommand\arraystretch{1.1} 
				\setlength{\aboverulesep}{0.5pt}
				\setlength{\belowrulesep}{0.5pt}
				\setlength{\tabcolsep}{0.5pt}
				
				\begin{tabular}{ll cccc}
					\toprule[0.8pt]
					
					\multicolumn{2}{c}{\multirow{2}{*}{\textbf{Method}}} & 
					\multicolumn{3}{c}{\textbf{ Average ASR (\%)$\uparrow$, ResNet-50}} & 
					\multirow{2}{*}{\textbf{Avg.}} \\
					\cmidrule(lr){3-5}
					
					& & \textbf{CNN} & \textbf{CNN Ensemble} & \textbf{Transformer} & \\
					\midrule
					
					& N/A & 22.22 & 8.48 & 8.01 & 15.24 \\
					& w/ DBO,w/o MAP & 30.82 & 10.11 & 8.40 & 20.04 \\
					& w/o DBO,w/ MAP & 29.24 & 12.67 & 9.67 & 20.21 \\
					
					\rowcolor{ourscore}
					\cellcolor{white} & \textbf{w/ DBO,w/ MAP} & \textbf{43.42} & \textbf{15.46} & \textbf{12.06} & \textbf{28.59} \\
					
					\cmidrule(lr){2-6}
					
					& AWT & 25.99 & 11.57 & 9.05 & 18.15 \\
					& w/ DBO,w/o MAP & 34.11 & 14.75 & 10.43 & 23.35 \\
					& w/o DBO,w/ MAP & 32.22 & 15.73 & 11.02 & 22.80 \\
					
					\rowcolor{ourscore}
					\cellcolor{white}\multirow{-8}{*}{\textbf{PGD}} & \textbf{w/ DBO,w/ MAP} & \textbf{38.88} & \textbf{18.93} & \textbf{12.09} & \textbf{27.19} \\
					
					\bottomrule[0.8pt]
				\end{tabular}
			\end{threeparttable}
		}
	\end{minipage}
	\vspace{-0.4cm} %
\end{wrapfigure}


\textbf{Efficiency vs. MAP Performance.}
As shown in Fig.~\ref{fig:loss-figure} (Right), the default Resize MAP already yields nearly $\times2$ ASR over PGD with negligible overhead, making it an efficient low-cost realization. Some stronger generative priors, such as SD-v1.5 and Tiny-SD, further improve ASR by about 4-5 points, but introduce higher latency.
This trend shows that MAP provides a scalable anchoring interface rather than a single fixed operator.


\noindent\textbf{Ablation of Hyperparameters ($T,\beta,\lambda$).}
As shown in Fig.~\ref{fig:hyperparameter} of \textit{\textbf{Appendix}~\ref{ablation: hyperparams}}, MABT is stable with respect to $\beta$ and $\lambda$, with limited ASR variation across broad ranges (e.g., $\lambda\in[10^{-3},10^{-1}]$).
In contrast, $T$ mainly controls the bilevel warm-up strength: too small $T$ under-aligns the initialization, while overly large $T$ may overfit the surrogate warm-up.
We set $T=3$ as the default trade-off between transferability and efficiency.  


%

\section{Conclusion}

We propose MABT to rectify the geometric disconnect in adversarial transferability. By synergizing MAP with a distributional bilevel framework, MABT encourages adversarial trajectories to remain closer to the shared semantic subspace. Furthermore, our Hessian-free solver enables low-overhead integration with diverse attacks, achieving strong performance across heterogeneous architectures. \noindent\textbf{\textit{Limitation.}} Current evaluation mainly focuses on image classification, while extending MABT to dense prediction tasks remains future work. This study may provide useful insights for more reliable robustness evaluation under black-box transfer settings.

\appendix

	\section{Appendix}

All the above experiments were conducted on a server equipped with an NVIDIA H100 GPU with 80GB memory.  The core implementation is provided in the \textbf{\texttt{Supplementary Material}} to facilitate reproducibility. The full code repository will be made public upon acceptance.

%
%

\subsection{Section Overview}
Below is a brief summary of the Appendix.

\begin{itemize}[leftmargin=2.2em,itemsep=0.2em,topsep=0.2em]
	\item \textbf{Method Details:} \textit{\textbf{Appendix} ~\ref{Method_Details}} provides practical details of MABT, including the local Gaussian instantiation of the surrogate distribution in DBO, finite-sample estimation of the expected transfer risk, and practical MAP realizations with their backward handling.
	
	\item \textbf{Runtime and Ablation Results:} \textit{\textbf{Appendix} ~\ref{Ablation_Results}} reports runtime-fair comparisons, ablations of different MAP priors, component studies of MAP and DBO, and hyperparameter sensitivity analyses.
	
	\item \textbf{Comparative Results:} \textit{\textbf{Appendix}~\ref{Comparative_Results}} provides extended empirical evaluations, including full results on additional attackers, VGG-19 surrogate results, targeted attack results, and defense scenarios.
	
	\item \textbf{Algorithm Analysis:} \textit{\textbf{Appendix}~\ref{sec:proof_convergence}} presents the assumptions and proof for the convergence result of the proposed gap-regularized solver.
\end{itemize}

\subsection{Method Details}\label{Method_Details}

\noindent\textbf{DBO Instantiation.} In the main text, the upper-level distribution $\pi(\boldsymbol{\theta}|\mathcal{S})$ is introduced to model local surrogate uncertainty. 
In practice, we do not assume access to the true distribution of unknown victim models. 
Instead, we instantiate $\pi(\boldsymbol{\theta}|\mathcal{S})$ as a local Gaussian approximation around a fine-tuned surrogate checkpoint.
Starting from the source surrogate $\mathcal{S}$, we first perform lightweight fine-tuning and then sample a small family of surrogate variants by adding Gaussian parameter perturbations around the resulting checkpoint.
The sampled variants are fixed during attack optimization, and the expectation in Eq.~\eqref{eq:rdist_def} is computed by finite-sample averaging:
\begin{equation}
	\mathcal{R}_{dist}(\boldsymbol{\delta};\pi)
	\approx
	\frac{1}{N}\sum_{i=1}^{N}
	\mathcal{L}\!\left(
	\mathcal{S}_{\boldsymbol{\theta}_i}
		(\mathbf{x}+\boldsymbol{\delta}),
	\mathbf{y}
	\right),
	\quad \boldsymbol{\theta}_i\sim\pi(\boldsymbol{\theta}|\mathcal{S}).
\end{equation}
We use $N=3$ surrogate variants in our implementation to keep the upper-level risk estimation efficient while capturing local surrogate uncertainty.
This finite-sample approximation serves as a practical realization of the distributional objective, where $\pi(\boldsymbol{\theta}|\mathcal{S})$ represents a surrogate-induced local hypothesis family rather than the exact distribution of unknown victim models. Practically, the surrogate is fine-tuned following MBA~\cite{li2023making} for 10 epochs at a batch size of 1024, utilizing the SGD optimizer with a learning rate of 0.001, momentum of 0.9, and weight decay of 5e-4.

\noindent\textbf{MAP Realizations and Practical Optimization.}
All MAP realizations are implemented through the same forward anchoring interface,
$\tilde{\mathbf{x}}=\mathcal{P}_{\mathcal{M}}[\mathbf{x}+\boldsymbol{\delta}]$,
where $\tilde{\mathbf{x}}$ denotes the anchored input used for surrogate evaluation.
They differ only in how the operator $\mathcal{P}_{\mathcal{M}}$ is instantiated.
We consider the following three categories:

\begin{itemize}[leftmargin=2.2em,itemsep=0.2em,topsep=0.2em]
	\item \textbf{Explicit structural priors.}	This category includes Resize, Bit-Depth, and FFT filtering.
	Resize applies bilinear downsampling followed by upsampling with a scale factor of $0.9$ to attenuate unstable high-frequency perturbations.
	Bit-Depth removes small-magnitude variations through low-bit quantization, and FFT filtering suppresses high-frequency components in the frequency domain.
	These lightweight operators provide efficient empirical anchoring with negligible computational overhead.
	\item \textbf{Discriminative denoising prior.} We instantiate this category with DnCNN~\cite{zhang2017beyond}.
	Given an input $\mathbf{u}$, DnCNN predicts a residual $r_{\psi}(\mathbf{u})$, and the anchored output is computed as $\mathbf{u}-r_{\psi}(\mathbf{u})$.
	This learned denoising prior suppresses noise-like perturbation components while preserving the main image structure.
	\item \textbf{Generative purification priors.}	This category includes SD-v1.5~\cite{rombach2022high}, Tiny-SD, and OFA-S.
	These diffusion-based operators map the perturbed input to a purified output, offering potentially stronger but more computationally expensive anchoring realizations.
\end{itemize}

For practical optimization, differentiable realizations are handled by their native autograd pipelines, while non-smooth or generative realizations use standard surrogate-gradient approximations for update estimation.
All variants share the same forward anchoring interface, whereas the theoretical analysis in the main text is stated for smooth composite settings satisfying the required assumptions.

\subsection{Ablation Results}\label{Ablation_Results}

\begin{table}[htbp]
	\centering
	\caption{Ablation study on the effectiveness of different MAP components based on PGD attacker.}
	\label{tab:ablation_study_MAP}
	\scalebox{0.92}{

		\begin{threeparttable}
			\fontsize{7.5pt}{8.5pt}\selectfont
			\renewcommand\arraystretch{1.2} 
			\setlength{\aboverulesep}{0.5pt}
			\setlength{\belowrulesep}{0.5pt}
			\setlength{\tabcolsep}{0.5pt}
			
			\begin{tabular}{ll cccccc ccc ccc c}
				\toprule[0.8pt]
				
				\multicolumn{2}{c}{\multirow{2}{*}{\textbf{Method}}} & 
				\multicolumn{6}{c}{\textbf{CNN}} & 
				\multicolumn{3}{c}{\textbf{CNN Ensemble}} & 
				\multicolumn{3}{c}{\textbf{Transformer}} &
				\multirow{2}{*}{\textbf{Avg.}} \\
				
				\cmidrule(lr){3-8} \cmidrule(lr){9-11} \cmidrule(lr){12-14}
				
				& & Inc-v3 & Inc-Res-v2 & DenseNet & MobileNet & PNASNet & SENet & $\text{Inc-v3}_{\text{ens3}}$ & $\text{Inc-v3}_{\text{ens4}}$ & $\text{IncRes-v2}_{\text{ens}}$ & Visformer-s &DeiT-s & ConViT-b & \\
				\midrule
				
				& N/A & 16.38 & 13.04 & 37.52 & 36.50 & 12.80 & 17.10 & 10.18 & 9.40 & 5.86 & 10.98 & 7.50 & 5.56 & 15.24 \\
												\cmidrule{2-15}
				& FFT & 32.42 & 23.00 & 64.06 & 64.22 & 24.34 & 28.66 & 17.96 & 15.18 & 8.98 & 14.44 & 10.90 & 7.12 & 25.94 \\
				
				& Bit-Depth & 34.68 & 25.50 & 69.78 & 70.78 & 25.38 & 33.76 & 19.08 & 15.44 & 9.50 & 20.28 & 11.08 & 7.50 & 28.56 \\
				
				& Resize & {34.96} & {24.58} & {70.00} & {69.22} & {27.84} & {33.94} & {19.72} & {16.38} & {10.28} & {17.02} & {11.54} & {7.62} & {28.59} \\
								\cmidrule{2-15}
				& DnCNN & 24.62 & 16.92 & 56.50 & 59.88 & 16.12 & 23.96 & 12.94 & 10.54 & 6.10 & 13.56 & 8.36 & 5.86 & 21.28 \\
								\cmidrule{2-15}
				& OFA-S & 31.20 & 23.08 & 57.32 & 59.48 & 22.04 & 26.16 & 19.76 & 17.82 & 11.06 & 16.72 & 11.72 & 8.46 & 25.40 \\
				
				& SD-v1.5 & 41.68 & 31.44 & 73.82 & 71.16 & 34.74 & 37.80 & 25.14 & 21.08 & 12.86 & 20.58 & 13.20 & 9.32 & 32.74 \\
				
				\rowcolor{ourscore} \multirow{-8}{*}{\textbf{MAP}}& Tiny-SD& \textbf{42.16} & \textbf{31.46} & \textbf{72.84} & \textbf{71.30} &\textbf{ 35.00} & \textbf{38.14} & \textbf{25.98} & \textbf{21.58} & \textbf{13.72} & \textbf{20.90} & \textbf{13.78} & \textbf{9.86} & \textbf{33.06} \\
				\bottomrule[0.8pt]
			\end{tabular}
		\end{threeparttable}
		
	}
\end{table}

\begin{table}[htbp]
	\centering
	\caption{Ablation study on the effectiveness of individual DBO and MAP components.}
	\label{tab:ablation_study_full}
	\scalebox{0.93}{

		\begin{threeparttable}
			\fontsize{7.pt}{8.pt}\selectfont
			\renewcommand\arraystretch{1.2} 
			\setlength{\aboverulesep}{0.5pt}
			\setlength{\belowrulesep}{0.5pt}
			\setlength{\tabcolsep}{0.5pt}
			
			\begin{tabular}{ll cccccc ccc ccc c}
				\toprule[0.8pt]
				
				\multicolumn{2}{c}{\multirow{2}{*}{\textbf{Method}}} & 
				\multicolumn{6}{c}{\textbf{CNN}} & 
				\multicolumn{3}{c}{\textbf{CNN Ensemble}} & 
				\multicolumn{3}{c}{\textbf{Transformer}} &
				\multirow{2}{*}{\textbf{Avg.}} \\
				
				\cmidrule(lr){3-8} \cmidrule(lr){9-11} \cmidrule(lr){12-14}
				
				& & Inc-v3 & Inc-Res-v2 & DenseNet & MobileNet & PNASNet & SENet & $\text{Inc-v3}_{\text{ens3}}$ & $\text{Inc-v3}_{\text{ens4}}$ & $\text{IncRes-v2}_{\text{ens}}$ & Visformer-s &DeiT-s & ConViT-b & \\
				\midrule
				
				& N/A & 16.38 & 13.04 & 37.52 & 36.50 & 12.80 & 17.10 & 10.18 & 9.40 & 5.86 & 10.98 & 7.50 & 5.56 & 15.24 \\
				& w/ DBO,w/o MAP & 22.64 & 15.76 & 52.58 & 57.24 & 14.52 & 22.18 & 13.02 & 10.80 & 6.50 & 12.26 & 7.92 & 5.02 & 20.04 \\
				& w/o DBO,w/ MAP & 22.78 & 18.28 & 47.94 & 43.84 & 20.70 & 21.92 & 14.80 & 14.30 & 8.90 & 11.98 & 10.02 & 7.02 & 20.21 \\
				\rowcolor{ourscore}
				\cellcolor{white} & \textbf{w/ DBO,w/ MAP} & \textbf{34.96} & \textbf{24.58} & \textbf{70.00} & \textbf{69.22} & \textbf{27.84} & \textbf{33.94} & \textbf{19.72} & \textbf{16.38} & \textbf{10.28} & \textbf{17.02} & \textbf{11.54} & \textbf{7.62} & \textbf{28.59} \\
				\cmidrule{2-15}
				
				& GHOST & 18.34 & 14.44 & 41.56 & 41.72 & 13.60 & 19.78 & 10.98 & 10.00 & 6.02 & 12.10 & 7.14 & 5.90 & 16.80 \\
				& w/ DBO,w/o MAP & 24.24 & 16.26 & 54.12 & 58.32 & 15.92 & 23.14 & 13.20 & 10.76 & 5.94 & 12.46 & 7.84 & 5.22 & 20.62 \\
				& w/o DBO,w/ MAP & 27.00 & 20.60 & 54.90 & 50.52 & 24.46 & 25.96 & 17.32 & 15.12 & 10.36 & 13.72 & 10.52 & 7.60 & 23.17 \\
				\rowcolor{ourscore}
				\cellcolor{white} & \textbf{w/ DBO,w/ MAP} & \textbf{37.38} & \textbf{27.52} & \textbf{74.66} & \textbf{72.58} & \textbf{30.84} & \textbf{36.72} & \textbf{21.18} & \textbf{17.66} & \textbf{10.50} & \textbf{18.24} & \textbf{11.92} & \textbf{7.72} & \textbf{30.58} \\
				\cmidrule{2-15}
				
				& MBA & 21.66 & 14.98 & 49.22 & 57.74 & 13.06 & 20.22 & 12.28 & 10.26 & 5.94 & 11.00 & 7.58 & 4.52 & 19.04 \\
				& w/ DBO,w/o MAP & 32.58 & 23.54 & 67.52 & 72.94 & 21.02 & 31.20 & 17.04 & 14.24 & 8.68 & 16.80 & 9.76 & 6.30 & 26.80 \\
				& w/o DBO,w/ MAP & 31.26 & 22.14 & 64.08 & 67.16 & 22.18 & 26.68 & 19.96 & 17.30 & 11.24 & 12.02 & 10.92 & 6.28 & 25.94 \\
				\rowcolor{ourscore}
				\cellcolor{white} & \textbf{w/ DBO,w/ MAP} & \textbf{44.20} & \textbf{32.52} & \textbf{78.76} & \textbf{81.22} & \textbf{32.60} & \textbf{39.20} & \textbf{27.12} & \textbf{22.44} & \textbf{14.54} & \textbf{19.08} & \textbf{13.80} & \textbf{8.36} & \textbf{34.49} \\
				\cmidrule{2-15}
				
				& AWT & 21.24 & 15.92 & 42.30 & 42.38 & 15.58 & 18.50 & 13.52 & 13.10 & 8.10 & 12.04 & 8.78 & 6.32 & 18.15 \\
				& w/ DBO,w/o MAP & 28.28 & 19.86 & 54.62 & 59.52 & 18.54 & 23.84 & 17.84 & 16.12 & 10.30 & 14.02 & 10.40 & 6.88 & 23.35 \\
				& w/o DBO,w/ MAP & 27.88 & 21.52 & 50.40 & 48.02 & 22.00 & 23.52 & 18.14 & 17.22 & 11.84 & 13.66 & 11.28 & 8.12 & 22.80 \\
				\rowcolor{ourscore}
				\cellcolor{white}\multirow{-16}{*}{\textbf{PGD}} & \textbf{w/ DBO,w/ MAP} & \textbf{33.94} & \textbf{24.58} & \textbf{59.56} & \textbf{61.92} & \textbf{24.68} & \textbf{28.60} & \textbf{22.52} & \textbf{20.76} & \textbf{13.52} & \textbf{15.52} & \textbf{12.58} & \textbf{8.12} & \textbf{27.19} \\
				
				\bottomrule[0.8pt]
			\end{tabular}
		\end{threeparttable}
	}
\end{table}

\begin{figure}[t]
	\centering
	\begin{minipage}[t]{0.49\linewidth}
		\centering
		\includegraphics[width=\linewidth]{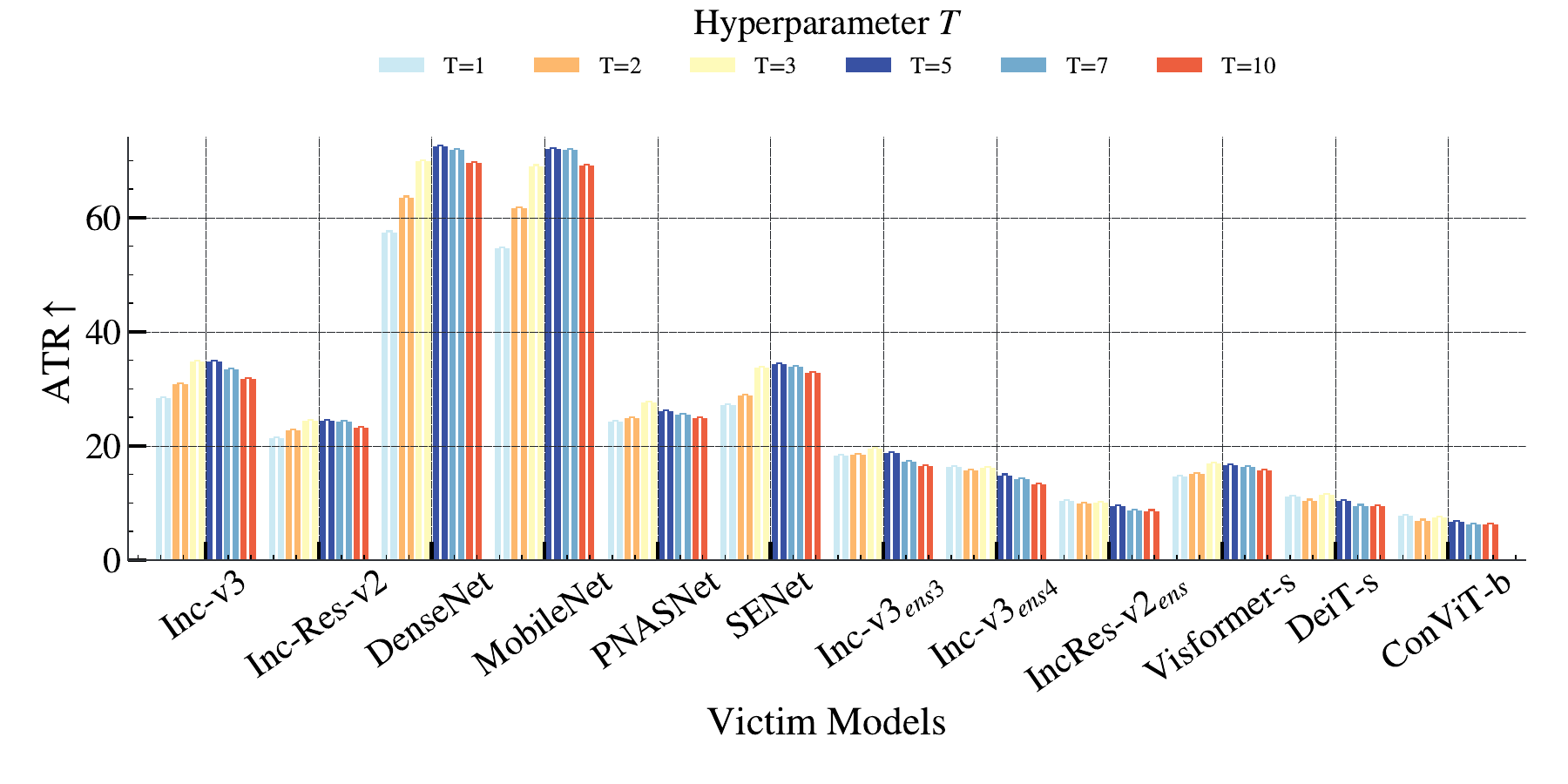}
		\vspace{0mm}
		\centerline{\small (a) $T$ on ResNet-50}
	\end{minipage}
	\hfill
	\begin{minipage}[t]{0.49\linewidth}
		\centering
		\includegraphics[width=\linewidth]{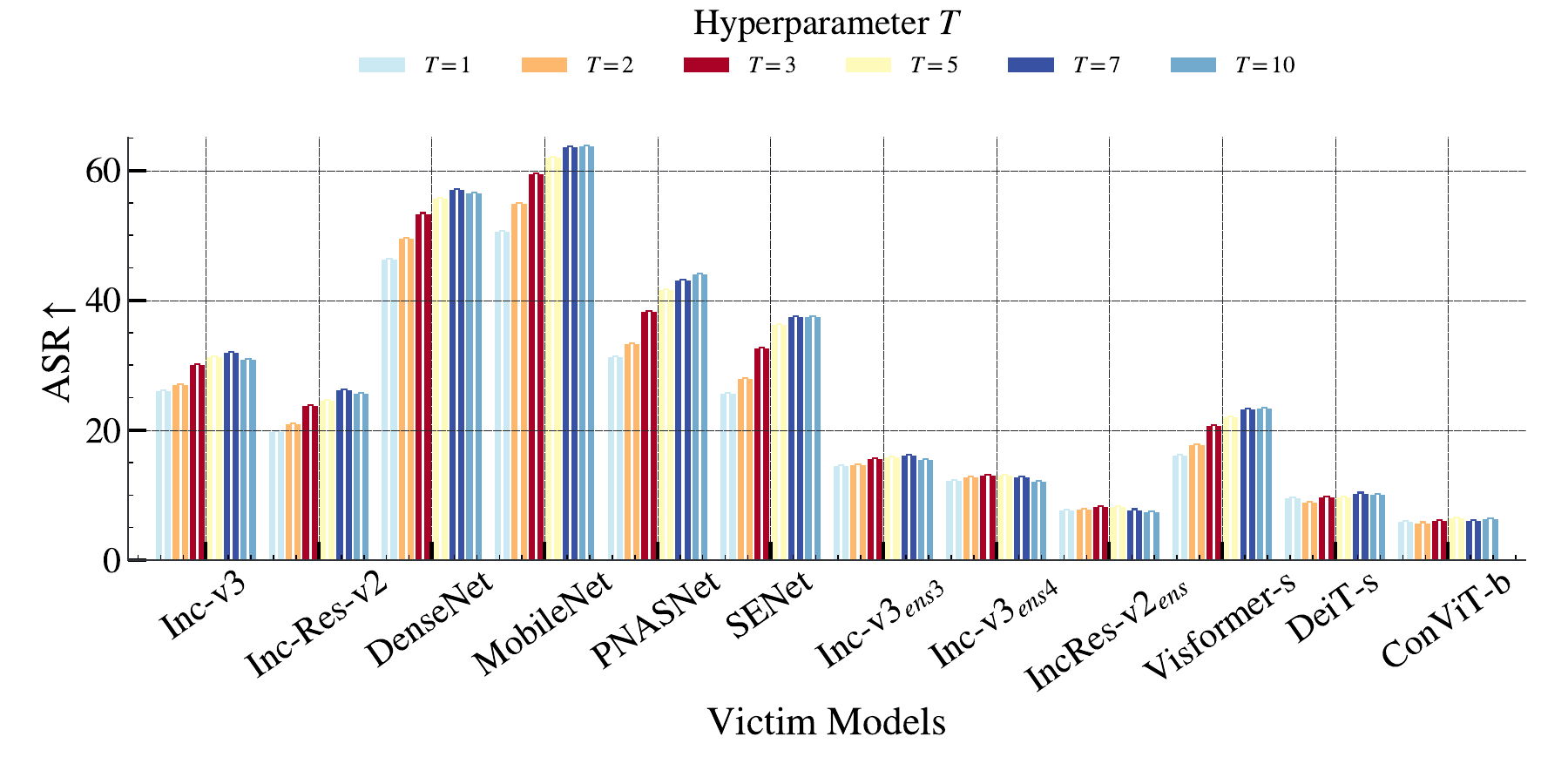}
		\vspace{0mm}
		\centerline{\small (b) $T$ on VGG-19}
	\end{minipage}
	
	\vspace{2mm}
	
	\begin{minipage}[t]{0.49\linewidth}
		\centering
		\includegraphics[width=\linewidth]{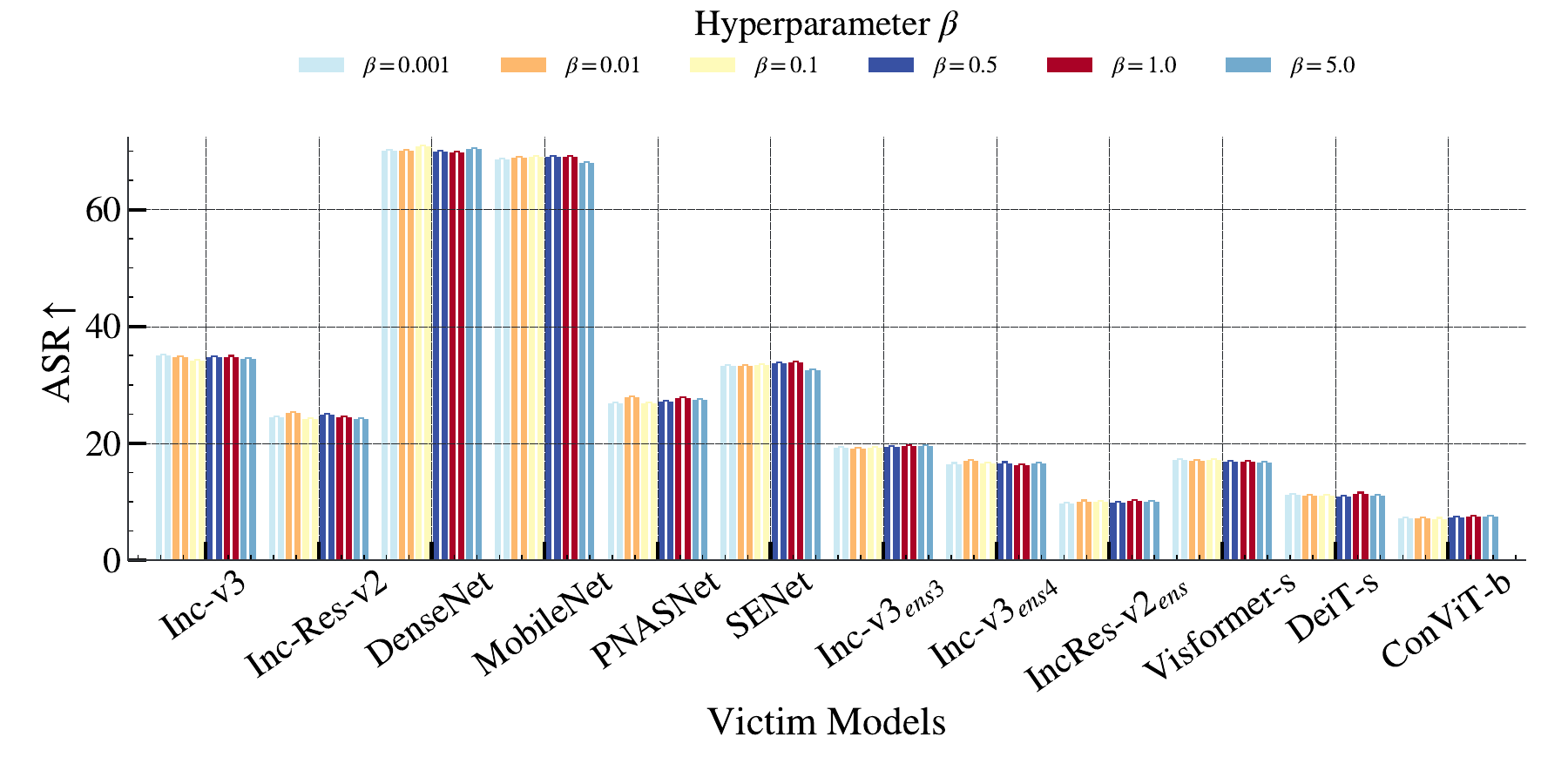}
		\vspace{0mm}
		\centerline{\small (c) $\beta$ on ResNet-50}
	\end{minipage}
	\hfill
	\begin{minipage}[t]{0.49\linewidth}
		\centering
		\includegraphics[width=\linewidth]{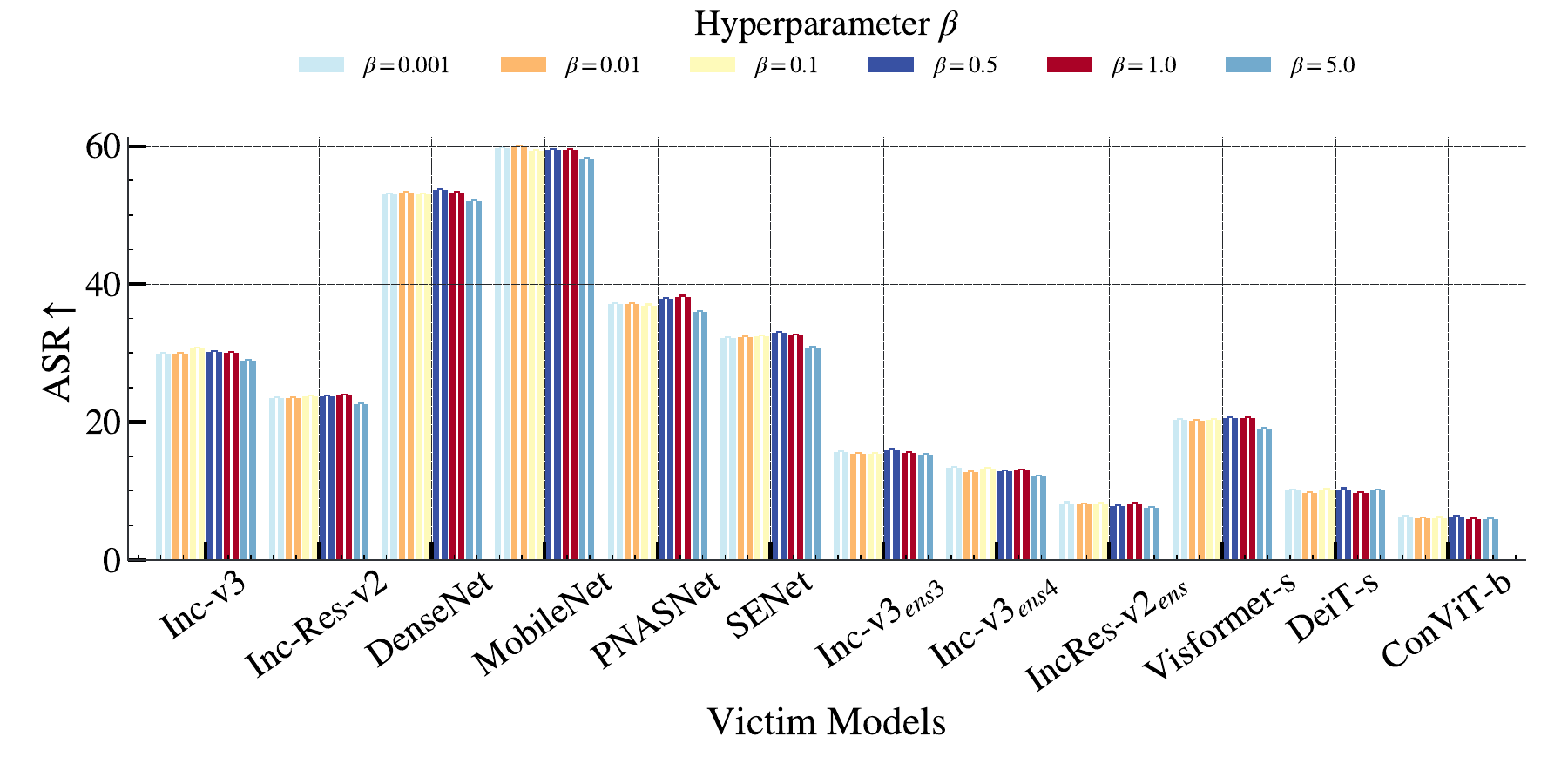}
		\vspace{0mm}
		\centerline{\small (d) $\beta$ on VGG-19}
	\end{minipage}
	
	\vspace{2mm}
	
	\begin{minipage}[t]{0.49\linewidth}
		\centering
		\includegraphics[width=\linewidth]{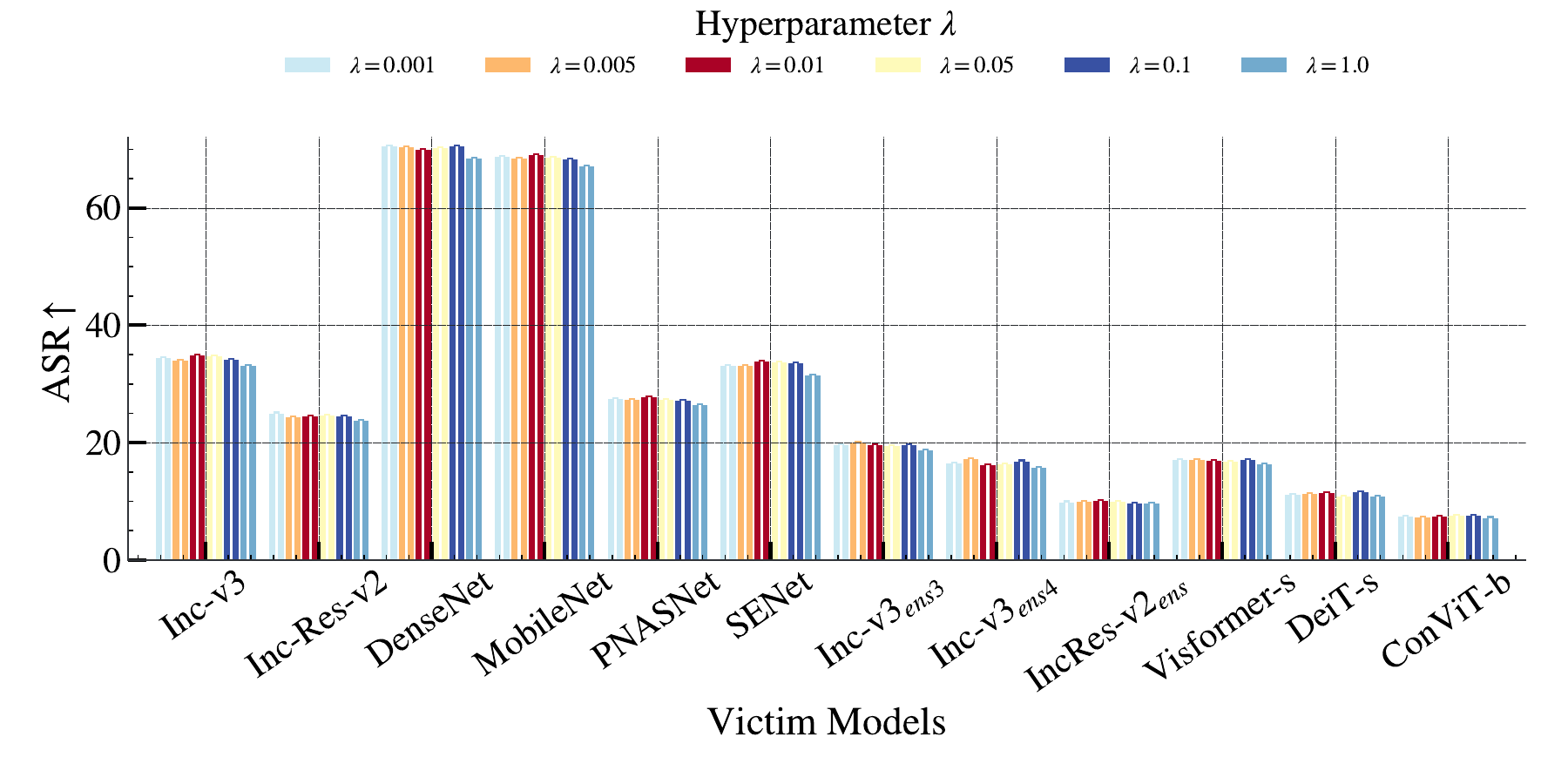}
		\vspace{0mm}
		\centerline{\small (e) $\lambda$ on ResNet-50}
	\end{minipage}
	\hfill
	\begin{minipage}[t]{0.49\linewidth}
		\centering
		\includegraphics[width=\linewidth]{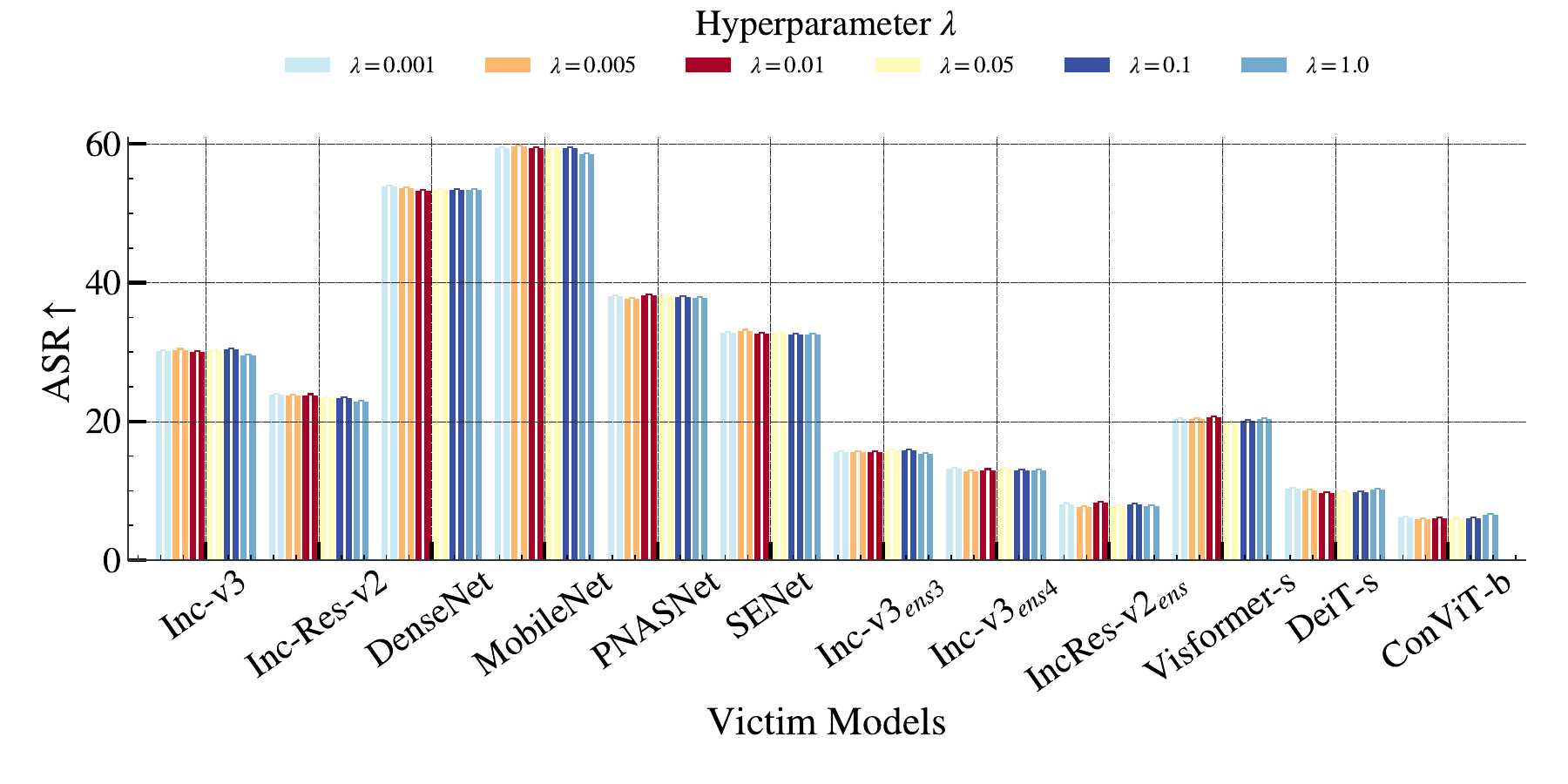}
		\vspace{0mm}
		\centerline{\small (f) $\lambda$ on VGG-19}
	\end{minipage}
	\caption{\textbf{Hyperparameter sensitivity on ResNet-50 and VGG-19 surrogates.}
		We evaluate the effects of the warm-up iteration $T$, lower-level step size $\beta$, and gap coefficient $\lambda$.}
	\label{fig:hyperparameter}
			\vspace {-0.2cm}
\end{figure}

\begin{table*}[t]
	\centering
\caption{Comparison of untargeted ASR (\%) by implementing 3 base attackers under MABT, including VMI, DI, and SI. We integrate and compare MABT with different method combinations. The best results are highlighted in \textbf{bold}.}
	\label{tab:main_results_full}
	\scalebox{0.97}{
		\begin{threeparttable}
			\fontsize{7.pt}{8.pt}\selectfont
			\renewcommand\arraystretch{1.2} 
			\setlength{\aboverulesep}{0.5pt}
			\setlength{\belowrulesep}{0.5pt}
			\setlength{\tabcolsep}{0.85pt}
			
			\begin{tabular}{ll cccccc ccc ccc c}
				\toprule[0.8pt]
				\multicolumn{15}{c}{\textbf{Untargeted Attack Scenario, ResNet-50 backbone, ASR (\%) $\uparrow$}} \\
				\cmidrule(lr){1-15} 
				
				\multicolumn{2}{c}{\multirow{2}{*}{\textbf{Method}}} & 
				\multicolumn{6}{c}{\textbf{CNN}} & 
				\multicolumn{3}{c}{\textbf{CNN Ensemble}} & 
				\multicolumn{3}{c}{\textbf{Transformer}} &
				\multirow{2}{*}{\textbf{Avg.}} \\
				
				\cmidrule(lr){3-8} \cmidrule(lr){9-11} \cmidrule(lr){12-14}
				
				& & Inc-v3 & Inc-Res-v2 & DenseNet & MobileNet & PNASNet & SENet & $\text{Inc-v3}_{\text{ens3}}$ & $\text{Inc-v3}_{\text{ens4}}$ & $\text{IncRes-v2}_{\text{ens}}$ & Visformer-s &DeiT-s & ConViT-b & \\
				\midrule
				& N/A & 36.98 & 31.44 & 62.64 & 59.84 & 33.42 & 39.52 & 23.08 & 20.34 & 14.10 & 25.96 & 16.78 & 13.18 & 31.44 \\
				\rowcolor{ourscore}
				\cellcolor{white} & \textbf{Ours} & \textbf{63.64} & \textbf{51.00} & \textbf{90.96} & \textbf{89.84} & \textbf{53.10} & \textbf{61.86} & \textbf{40.50} & \textbf{34.10} & \textbf{22.98} & \textbf{37.46} & \textbf{23.44} & \textbf{16.88} & \textbf{48.81} \\
				
				& GHOST & 39.58 & 32.42 & 67.82 & 64.70 & 34.50 & 42.10 & 24.38 & 21.74 & 14.34 & 26.64 & 17.38 & 12.84 & 33.20 \\
				\rowcolor{ourscore}
				\cellcolor{white} & \textbf{Ours} & \textbf{66.22} & \textbf{53.82} & \textbf{92.64} & \textbf{91.62} & \textbf{56.46} & \textbf{65.62} & \textbf{41.76} & \textbf{35.72} & \textbf{24.20} & \textbf{41.26} & \textbf{26.08} & \textbf{18.48} & \textbf{51.16} \\
				
				& MBA & 44.16 & 32.06 & 76.32 & 80.92 & 30.80 & 40.30 & 26.00 & 22.04 & 14.08 & 23.52 & 16.10 & 10.40 & 34.73 \\
				\rowcolor{ourscore}
				\cellcolor{white} & \textbf{Ours} & \textbf{66.16} & \textbf{53.96} & \textbf{92.70} & \textbf{93.74} & \textbf{54.06} & \textbf{62.80} & \textbf{43.28} & \textbf{36.36} & \textbf{24.68} & \textbf{40.40} & \textbf{25.38} & \textbf{17.52} & \textbf{50.92} \\
				
				& AWT & 39.90 & 32.40 & 63.78 & 62.12 & 33.44 & 35.52 & 27.70 & 25.74 & 18.02 & 23.02 & 17.64 & 13.12 & 32.70 \\
				\rowcolor{ourscore}
				\cellcolor{white}\multirow{-8}{*}{\textbf{VMI}} & \textbf{Ours} & \textbf{57.86} & \textbf{46.52} & \textbf{84.48} & \textbf{85.10} & \textbf{46.86} & \textbf{52.78} & \textbf{40.36} & \textbf{36.38} & \textbf{25.08} & \textbf{33.38} & \textbf{24.58} & \textbf{17.10} & \textbf{45.87} \\
				\midrule
				& N/A & 37.12 & 31.96 & 68.84 & 64.00 & 36.54 & 40.00 & 22.96 & 20.10 & 13.16 & 21.76 & 13.58 & 10.48 & 31.71 \\
				\rowcolor{ourscore}
				\cellcolor{white} & \textbf{Ours} & \textbf{56.72} & \textbf{44.88} & \textbf{87.70} & \textbf{84.24} & \textbf{52.94} & \textbf{53.82} & \textbf{37.30} & \textbf{30.60} & \textbf{20.82} & \textbf{27.10} & \textbf{17.46} & \textbf{11.76} & \textbf{43.78} \\
				
				& GHOST & 38.66 & 30.66 & 70.12 & 65.38 & 35.30 & 39.12 & 23.10 & 19.42 & 12.76 & 20.72 & 13.12 & 10.38 & 31.56 \\
				\rowcolor{ourscore}
				\cellcolor{white} & \textbf{Ours} & \textbf{57.12} & \textbf{45.24} & \textbf{87.90} & \textbf{85.58} & \textbf{51.46} & \textbf{53.64} & \textbf{36.60} & \textbf{30.64} & \textbf{20.02} & \textbf{26.90} & \textbf{17.08} & \textbf{11.88} & \textbf{43.67} \\
				
				& MBA & 44.84 & 32.40 & 78.94 & 81.34 & 32.28 & 38.90 & 27.16 & 23.04 & 15.08 & 20.08 & 14.76 & 8.86 & 34.81 \\
				\rowcolor{ourscore}
				\cellcolor{white} & \textbf{Ours} & \textbf{60.56} & \textbf{47.46} & \textbf{89.48} & \textbf{89.78} & \textbf{49.64} & \textbf{52.56} & \textbf{42.42} & \textbf{37.94} & \textbf{26.22} & \textbf{26.90} & \textbf{19.94} & \textbf{12.30} & \textbf{46.27} \\
				
				& AWT & 38.98 & 31.42 & 66.24 & 62.38 & 33.54 & 34.18 & 26.96 & 24.14 & 16.76 & 20.56 & 14.70 & 10.50 & 31.70 \\
				\rowcolor{ourscore}
				\cellcolor{white}\multirow{-8}{*}{\textbf{DI}} & \textbf{Ours} & \textbf{46.06} & \textbf{35.06} & \textbf{72.78} & \textbf{71.98} & \textbf{36.32} & \textbf{37.64} & \textbf{32.46} & \textbf{30.42} & \textbf{21.02} & \textbf{20.42} & \textbf{15.56} & \textbf{11.06} & \textbf{35.90} \\
				\midrule
				& N/A & 22.12 & 17.70 & 48.76 & 46.46 & 17.66 & 23.38 & 13.14 & 11.50 & 7.02 & 15.32 & 9.20 & 7.14 & 19.95 \\
				\rowcolor{ourscore}
				\cellcolor{white} & \textbf{Ours} & \textbf{45.36} & \textbf{33.20} & \textbf{80.16} & \textbf{76.48} & \textbf{35.92} & \textbf{41.38} & \textbf{26.96} & \textbf{21.70} & \textbf{13.96} & \textbf{21.34} & \textbf{13.68} & \textbf{9.44} & \textbf{34.97} \\
				
				& GHOST & 23.82 & 18.38 & 53.70 & 51.80 & 19.32 & 25.80 & 14.52 & 11.88 & 7.60 & 16.18 & 9.20 & 6.88 & 21.59 \\
				\rowcolor{ourscore}
				\cellcolor{white} & \textbf{Ours} & \textbf{48.04} & \textbf{35.98} & \textbf{82.92} & \textbf{79.48} & \textbf{38.58} & \textbf{43.86} & \textbf{28.94} & \textbf{23.84} & \textbf{14.96} & \textbf{23.36} & \textbf{15.08} & \textbf{9.74} & \textbf{37.07} \\
				
				& MBA & 24.70 & 18.10 & 51.82 & 59.76 & 14.10 & 22.66 & 13.98 & 11.06 & 7.04 & 12.16 & 7.60 & 4.76 & 20.65 \\
				\rowcolor{ourscore}
				\cellcolor{white} & \textbf{Ours} & \textbf{48.20} & \textbf{36.12} & \textbf{79.10} & \textbf{81.22} & \textbf{34.56} & \textbf{40.76} & \textbf{31.10} & \textbf{26.46} & \textbf{18.26} & \textbf{20.34} & \textbf{15.20} & \textbf{8.66} & \textbf{36.67} \\
				
				& AWT & 26.48 & 19.66 & 50.60 & 49.24 & 19.42 & 22.84 & 17.68 & 16.42 & 10.76 & 14.84 & 10.96 & 7.70 & 22.22 \\
				\rowcolor{ourscore}
				\cellcolor{white}\multirow{-8}{*}{\textbf{SI}} & \textbf{Ours} & \textbf{41.88} & \textbf{31.50} & \textbf{70.02} & \textbf{69.82} & \textbf{31.84} & \textbf{34.88} & \textbf{28.90} & \textbf{26.70} & \textbf{18.20} & \textbf{19.94} & \textbf{15.10} & \textbf{10.46} & \textbf{33.27} \\
				
				\bottomrule[0.8pt]
			\end{tabular}
		\end{threeparttable}
	}
	\vspace{-0.4cm}
\end{table*}

\begin{table*}[t]
	\centering
	\caption{Comparison of untargeted ASR (\%) using a VGG-19 surrogate. We implement 4 base attackers along with 3 ensembles  attacks. The best results in each category are highlighted in \textbf{bold}.}
	\label{tab:vgg_results}
	\scalebox{0.97}{
		\begin{threeparttable}
			\fontsize{7.pt}{8.pt}\selectfont
			\renewcommand\arraystretch{1.2} 
			\setlength{\aboverulesep}{0.5pt}
			\setlength{\belowrulesep}{0.5pt}
			\setlength{\tabcolsep}{.85pt}
			
			\begin{tabular}{ll cccccc ccc ccc c}
				\toprule[0.8pt]
				\multicolumn{15}{c}{\textbf{Untargeted Attack on ImageNet, VGGNet-19 backbone, ASR (\%) $\uparrow$}} \\
				\cmidrule(lr){1-15} 
				
				\multicolumn{2}{c}{\multirow{2}{*}{\textbf{Method}}} & 
				\multicolumn{6}{c}{\textbf{CNN}} & 
				\multicolumn{3}{c}{\textbf{CNN Ensemble}} & 
				\multicolumn{3}{c}{\textbf{Transformer}} &
				\multirow{2}{*}{\textbf{Avg.}} \\
				
				\cmidrule(lr){3-8} \cmidrule(lr){9-11} \cmidrule(lr){12-14}
				
				& & Inc-v3 & Inc-Res-v2 & DenseNet & MobileNet & PNASNet & SENet & $\text{Inc-v3}_{\text{ens3}}$ & $\text{Inc-v3}_{\text{ens4}}$ & $\text{IncRes-v2}_{\text{ens}}$ & Visformer-s &DeiT-s & ConViT-b & \\
				\midrule
				
				& N/A & 12.84 & 9.54 & 25.50 & 32.44 & 13.06 & 13.04 & 7.96 & 7.46 & 4.18 & 9.52 & 5.10 & 3.08 & 11.98 \\
				
				\rowcolor{ourscore}
				\cellcolor{white} & \textbf{Ours} & \textbf{29.74} & \textbf{23.50} & \textbf{52.56} & \textbf{58.34} & \textbf{37.58} & \textbf{32.70} & \textbf{15.38} & \textbf{12.92} & \textbf{7.94} & \textbf{19.84} & \textbf{10.28} & \textbf{6.28} & \textbf{25.59} \\
				
				& GHOST & 12.60 & 9.64 & 25.18 & 32.52 & 13.46 & 13.78 & 7.96 & 7.42 & 4.10 & 9.54 & 5.30 & 3.02 & 12.04 \\
				
				\rowcolor{ourscore}
				\cellcolor{white} & \textbf{Ours} & \textbf{29.94} & \textbf{23.22} & \textbf{52.50} & \textbf{58.46} & \textbf{37.38} & \textbf{32.02} & \textbf{15.26} & \textbf{12.94} & \textbf{8.16} & \textbf{19.88} & \textbf{10.14} & \textbf{6.42} & \textbf{25.53} \\
				
				& MBA & 25.26 & 19.96 & 49.56 & 58.96 & 29.78 & 28.40 & 13.16 & 11.60 & 6.78 & 20.46 & 8.90 & 5.18 & 23.17 \\
				
				\rowcolor{ourscore}
				\cellcolor{white} & \textbf{Ours} & \textbf{46.38} & \textbf{38.46} & \textbf{72.78} & \textbf{76.90} & \textbf{58.46} & \textbf{49.20} & \textbf{25.64} & \textbf{20.66} & \textbf{13.20} & \textbf{32.96} & \textbf{15.46} & \textbf{9.98} & \textbf{38.34} \\
				
				& AWT & 16.24 & 11.86 & 28.86 & 34.70 & 15.04 & 12.56 & 10.32 & 10.00 & 5.76 & 8.90 & 5.32 & 3.34 & 13.58 \\
				
				\rowcolor{ourscore}
				\cellcolor{white}\multirow{-8}{*}{\textbf{PGD}} & \textbf{Ours} & \textbf{26.20} & \textbf{18.94} & \textbf{42.46} & \textbf{50.36} & \textbf{24.76} & \textbf{22.48} & \textbf{14.94} & \textbf{13.44} & \textbf{8.22} & \textbf{14.14} & \textbf{9.18} & \textbf{5.14} & \textbf{20.86} \\
				\midrule
				
				& N/A & 19.72 & 15.46 & 37.08 & 43.34 & 21.34 & 23.16 & 11.62 & 10.12 & 5.86 & 14.50 & 8.26 & 4.94 & 17.95 \\
				\rowcolor{ourscore}
				\cellcolor{white} & \textbf{Ours} & \textbf{38.80} & \textbf{30.30} & \textbf{62.22} & \textbf{67.34} & \textbf{46.24} & \textbf{42.72} & \textbf{19.56} & \textbf{16.28} & \textbf{10.46} & \textbf{27.12} & \textbf{14.16} & \textbf{9.12} & \textbf{32.03} \\
				
				& GHOST & 19.84 & 15.48 & 36.46 & 43.14 & 21.48 & 23.18 & 11.50 & 10.04 & 6.20 & 14.22 & 8.42 & 5.00 & 17.91 \\
				\rowcolor{ourscore}
				\cellcolor{white} & \textbf{Ours} & \textbf{38.80} & \textbf{30.30} & \textbf{62.66} & \textbf{67.28} & \textbf{46.46} & \textbf{42.60} & \textbf{19.80} & \textbf{16.18} & \textbf{10.72} & \textbf{27.18} & \textbf{14.28} & \textbf{8.96} & \textbf{32.10} \\
				
				& MBA & 43.28 & 34.32 & 68.86 & 74.34 & 47.50 & 44.82 & 22.80 & 19.50 & 12.20 & 32.00 & 16.00 & 10.10 & 35.48 \\
				\rowcolor{ourscore}
				\cellcolor{white} & \textbf{Ours} & \textbf{56.96} & \textbf{47.14} & \textbf{79.76} & \textbf{82.26} & \textbf{65.44} & \textbf{59.20} & \textbf{32.16} & \textbf{25.94} & \textbf{16.64} & \textbf{41.10} & \textbf{20.96} & \textbf{14.38} & \textbf{45.16} \\
				
				& AWT & 25.72 & 18.34 & 40.64 & 46.12 & 23.66 & 21.08 & 15.36 & 14.58 & 9.46 & 13.08 & 8.32 & 5.08 & 20.12 \\
				\rowcolor{ourscore}
				\cellcolor{white} \multirow{-8}{*}{\textbf{MI}} & \textbf{Ours} & \textbf{37.68} & \textbf{28.22} & \textbf{56.94} & \textbf{61.96} & \textbf{37.40} & \textbf{34.40} & \textbf{20.84} & \textbf{18.68} & \textbf{11.58} & \textbf{23.84} & \textbf{13.62} & \textbf{8.40} & \textbf{29.46} \\
				\midrule
				& N/A & 14.12 & 10.88 & 28.50 & 35.80 & 17.28 & 17.70 & 8.78 & 8.08 & 4.90 & 10.96 & 6.22 & 3.64 & 13.91 \\
				\rowcolor{ourscore}
				\cellcolor{white} & \textbf{Ours} & \textbf{28.60} & \textbf{21.44} & \textbf{49.48} & \textbf{55.28} & \textbf{36.28} & \textbf{30.60} & \textbf{15.32} & \textbf{12.86} & \textbf{8.26} & \textbf{17.72} & \textbf{10.42} & \textbf{6.00} & \textbf{24.36} \\
				
				& GHOST & 13.92 & 10.64 & 28.38 & 35.78 & 17.30 & 17.38 & 9.02 & 7.84 & 4.88 & 10.42 & 6.14 & 3.46 & 13.76 \\
				\rowcolor{ourscore}
				\cellcolor{white} & \textbf{Ours} & \textbf{28.16} & \textbf{21.88} & \textbf{49.72} & \textbf{55.28} & \textbf{36.72} & \textbf{30.96} & \textbf{15.24} & \textbf{13.34} & \textbf{8.58} & \textbf{17.98} & \textbf{10.68} & \textbf{6.06} & \textbf{24.55} \\
				
				& MBA & 27.50 & 22.24 & 52.56 & 61.38 & 37.96 & 34.52 & 15.18 & 12.50 & 7.74 & 21.18 & 10.50 & 6.76 & 25.84 \\
				\rowcolor{ourscore}
				\cellcolor{white} & \textbf{Ours} & \textbf{41.82} & \textbf{34.08} & \textbf{68.04} & \textbf{72.16} & \textbf{55.44} & \textbf{45.20} & \textbf{23.32} & \textbf{19.92} & \textbf{12.52} & \textbf{28.28} & \textbf{15.12} & \textbf{9.46} & \textbf{35.45} \\
				
				& AWT & 19.46 & 13.42 & 33.04 & 39.14 & 17.80 & 15.28 & 11.56 & 11.42 & 6.92 & 8.76 & 6.74 & 3.70 & 15.60 \\
				\rowcolor{ourscore}
				\cellcolor{white}\multirow{-8}{*}{\textbf{TI}} & \textbf{Ours} & \textbf{27.40} & \textbf{19.12} & \textbf{43.62} & \textbf{50.22} & \textbf{25.30} & \textbf{22.42} & \textbf{16.10} & \textbf{14.70} & \textbf{9.34} & \textbf{13.14} & \textbf{9.64} & \textbf{5.36} & \textbf{21.36} \\
				\midrule
				
				& N/A & 25.66 & 16.88 & 42.58 & 50.36 & 22.26 & 19.54 & 14.14 & 13.60 & 7.90 & 13.16 & 8.72 & 5.42 & 20.02 \\
				\rowcolor{ourscore}
				\cellcolor{white} & \textbf{Ours} & \textbf{53.76} & \textbf{44.50} & \textbf{75.06} & \textbf{77.14} & \textbf{61.80} & \textbf{53.76} & \textbf{32.72} & \textbf{28.60} & \textbf{20.00} & \textbf{38.86} & \textbf{21.38} & \textbf{14.78} & \textbf{43.53} \\
				
				& GHOST & 25.50 & 17.00 & 42.76 & 49.96 & 21.62 & 19.52 & 14.80 & 13.76 & 7.90 & 13.24 & 8.78 & 5.14 & 20.00 \\
				\rowcolor{ourscore}
				\cellcolor{white} & \textbf{Ours} & \textbf{53.80} & \textbf{43.86} & \textbf{75.04} & \textbf{77.04} & \textbf{60.98} & \textbf{53.74} & \textbf{32.84} & \textbf{28.14} & \textbf{19.70} & \textbf{38.76} & \textbf{21.14} & \textbf{15.18} & \textbf{43.35} \\
				
				& MBA & 41.92 & 29.26 & 65.90 & 74.42 & 40.84 & 35.80 & 23.04 & 21.96 & 13.44 & 25.16 & 13.80 & 8.42 & 32.83 \\
				\rowcolor{ourscore}
				\cellcolor{white} & \textbf{Ours} & \textbf{70.50} & \textbf{60.52} & \textbf{88.64} & \textbf{89.40} & \textbf{78.30} & \textbf{69.90} & \textbf{47.60} & \textbf{43.16} & \textbf{31.70} & \textbf{53.66} & \textbf{30.62} & \textbf{22.46} & \textbf{57.21} \\
				
				& AWT & 30.42 & 20.94 & 49.02 & 56.40 & 28.10 & 22.80 & 19.24 & 18.56 & 12.00 & 15.62 & 10.42 & 6.34 & 24.16 \\
				\rowcolor{ourscore}
				\cellcolor{white} \multirow{-8}{*}{\textbf{GAA}} & \textbf{Ours} & \textbf{39.68} & \textbf{28.90} & \textbf{57.16} & \textbf{61.90} & \textbf{37.36} & \textbf{31.76} & \textbf{25.34} & \textbf{23.56} & \textbf{15.80} & \textbf{20.72} & \textbf{13.90} & \textbf{8.62} & \textbf{30.39} \\
				
				\bottomrule[0.8pt]
			\end{tabular}
		\end{threeparttable}
	}
	\vspace{-0.5cm}
\end{table*}

\subsubsection{Ablation of Different MAP Implementations}

To empirically validate the flexibility of MABT, we instantiate the MAP with seven practical realizations covering the three categories described above.
As visualized in the bubble chart in Fig.~\ref{fig:loss-figure} (Right) and reported in Tab.~\ref{tab:ablation_study_MAP}, MABT consistently improves transferability across different MAP choices, while exhibiting distinct efficiency--performance trade-offs.

\noindent\textbf{Performance vs. Efficiency Trade-off.}
Among generative priors, SD-v1.5 and Tiny-SD achieve the highest ASR, reaching 32.74\% and 33.06\% average ASR, respectively.
However, these stronger purification-based realizations also introduce substantially higher latency and memory consumption.
OFA-S, despite being a compact diffusion variant, attains 25.40\% average ASR and does not outperform the simpler Resize prior.
This suggests that generative priors can provide stronger anchoring effects, but their advantage is not uniform once model compression and computational cost are considered.

In contrast, Resize achieves a competitive 28.59\% average ASR with negligible overhead.
Although it is not the strongest MAP realization in terms of ASR, it provides the most favorable efficiency--performance balance among the tested variants.
Therefore, we adopt Resize as the default MAP realization in the main experiments.
Overall, these results show that MABT is not tied to a specific MAP operator, but provides a flexible anchoring interface that can accommodate lightweight structural priors, discriminative denoisers, and generative purification operators.

\subsubsection{Effectiveness of MAP and DBO} 

The full results in Tab.~\ref{tab:ablation_study_full} dissect the individual contributions of MAP and  DBO across various base attackers. Specifically, MAP suppresses surrogate-specific off-manifold noise to rectify the adversarial trajectory, while DBO learns a geometry-aligned initialization that elevates the starting point of the feature shift. Crucially, the joint optimization of MAP and DBO achieves the best performance, outperforming individual components by a significant margin. This synergy compels the adversarial trajectory to converge on shared semantic features.

\subsubsection{ Hyperparameter Analysis} \label{ablation: hyperparams}

Fig.~\ref{fig:hyperparameter} visualizes the sensitivity of $T$, $\beta$, and $\lambda$ on ResNet-50 and VGG-19 surrogates. MABT is stable to $\beta$ and $\lambda$ over broad ranges, suggesting that the gap-regularized update is not sensitive to their exact values. By contrast, $T$ controls the bilevel warm-up strength: small $T$ may under-align the initialization, whereas overly large $T$ may overfit the warm-up to surrogate-specific directions. The consistent trends across both backbones support the default setting $T=3$, $\beta=1.0$, and $\lambda=0.01$.

We note that $\sigma$ determines the probing point for estimating $\nabla\mathcal{G}$, rather than serving as an explicit update coefficient.
Because the final meta update applies an element-wise sign operation to $\lambda^{-1}v-\nabla\mathcal{G}$, the gap term can still provide coordinate-level rectification where the primary risk-gradient signal is weak, sign-unstable, or close to a sign transition.
Its practical effect is reflected in the component ablation and convergence behavior.


\begin{table*}[t]
	\centering
	\caption{Comparative results of targeted ASR against 12 victim architectures.}
	\label{tab:target_results}
	
	\scalebox{0.97}{
		\begin{threeparttable}
			\fontsize{7.pt}{8.pt}\selectfont
			\renewcommand\arraystretch{1.2} 
			\setlength{\aboverulesep}{0.5pt}
			\setlength{\belowrulesep}{0.5pt}
			\setlength{\tabcolsep}{0.85pt}
			
			\begin{tabular}{ll cccccc ccc ccc c}
				\toprule[0.8pt]
				\multicolumn{15}{c}{\textbf{Targeted Attack on ImageNet, ResNet-50 backbone, ASR (\%) $\uparrow$}} \\
				\cmidrule(lr){1-15} 
				
				\multicolumn{2}{c}{\multirow{2}{*}{\textbf{Method}}} & 
				\multicolumn{6}{c}{\textbf{CNN}} & 
				\multicolumn{3}{c}{\textbf{CNN Ensemble}} & 
				\multicolumn{3}{c}{\textbf{Transformer}} &
				\multirow{2}{*}{\textbf{Avg.}} \\
				
				\cmidrule(lr){3-8} \cmidrule(lr){9-11} \cmidrule(lr){12-14}
				
				& & Inc-v3 & Inc-Res-v2 & DenseNet & MobileNet & PNASNet & SENet & $\text{Inc-v3}_{\text{ens3}}$ & $\text{Inc-v3}_{\text{ens4}}$ & $\text{IncRes-v2}_{\text{ens}}$ & Visformer-s &DeiT-s & ConViT-b & \\
				\midrule
				
				& N/A & 0.02 & 0.00 & 0.04 & 0.08 & 0.00 & 0.00 & 0.00 & 0.00 & 0.00 & 0.00 & 0.00 & 0.00 & 0.01 \\
				\rowcolor{ourscore}
				\cellcolor{white} & \textbf{Ours} & \textbf{0.08} & \textbf{0.10} & \textbf{1.32} & \textbf{0.74} & \textbf{0.22} & \textbf{0.30} & \textbf{0.06} & \textbf{0.02} & \textbf{0.00} & \textbf{0.04} & \textbf{0.02} & \textbf{0.00} & \textbf{0.24} \\
				
				& GHOST & 0.02 & 0.00 & 0.14 & 0.08 & 0.00 & 0.04 & 0.00 & 0.00 & 0.00 & 0.00 & 0.00 & 0.00 & 0.02 \\
				\rowcolor{ourscore}
				\cellcolor{white} & \textbf{Ours} & \textbf{0.32} & \textbf{0.22} & \textbf{2.94} & \textbf{1.30} & \textbf{0.50} & \textbf{0.50} & \textbf{0.06} & \textbf{0.00} & \textbf{0.00} & \textbf{0.14} & \textbf{0.02} & \textbf{0.00} & \textbf{0.50} \\
				
				& MBA & 0.10 & 0.06 & 1.04 & 0.98 & 0.20 & 0.22 & 0.02 & 0.00 & 0.00 & 0.06 & 0.02 & 0.00 & 0.23 \\
				\rowcolor{ourscore}
				\cellcolor{white} & \textbf{Ours} & \textbf{0.98} & \textbf{0.70} & \textbf{7.50} & \textbf{5.38} & \textbf{1.10} & \textbf{1.30} & \textbf{0.28} & \textbf{0.22} & \textbf{0.12} & \textbf{0.44} & \textbf{0.10} & \textbf{0.06} & \textbf{1.52} \\
				
				& AWT & 0.00 & 0.00 & 0.06 & 0.04 & 0.02 & 0.00 & 0.00 & 0.00 & 0.00 & 0.00 & 0.00 & 0.00 & 0.01 \\
				\rowcolor{ourscore}
				\cellcolor{white}\multirow{-8}{*}{\textbf{PGD}} & \textbf{Ours} & \textbf{0.24} & \textbf{0.04} & \textbf{1.62} & \textbf{0.90} & \textbf{0.16} & \textbf{0.18} & \textbf{0.04} & \textbf{0.00} & \textbf{0.00} & \textbf{0.06} & \textbf{0.00} & \textbf{0.00} & \textbf{0.27} \\
				\midrule
				
				& N/A & 0.04 & 0.02 & 0.14 & 0.14 & 0.04 & 0.04 & 0.02 & 0.02 & 0.00 & 0.02 & 0.00 & 0.00 & 0.04 \\
				\rowcolor{ourscore}
				\cellcolor{white} & \textbf{Ours} & \textbf{0.74} & \textbf{0.58} & \textbf{4.92} & \textbf{3.06} & \textbf{0.82} & \textbf{1.04} & \textbf{0.22} & \textbf{0.10} & \textbf{0.02} & \textbf{0.38} & \textbf{0.14} & \textbf{0.04} & \textbf{1.01} \\
				
				& GHOST & 0.06 & 0.02 & 0.40 & 0.24 & 0.06 & 0.12 & 0.02 & 0.00 & 0.00 & 0.02 & 0.00 & 0.00 & 0.08 \\
				\rowcolor{ourscore}
				\cellcolor{white} & \textbf{Ours} & \textbf{1.22} & \textbf{0.68} & \textbf{7.82} & \textbf{4.46} & \textbf{1.42} & \textbf{1.74} & \textbf{0.40} & \textbf{0.12} & \textbf{0.06} & \textbf{0.54} & \textbf{0.18} & \textbf{0.06} & \textbf{1.56} \\
				
				& MBA & 0.32 & 0.14 & 2.34 & 1.78 & 0.46 & 0.44 & 0.06 & 0.04 & 0.00 & 0.20 & 0.04 & 0.02 & 0.49 \\
				\rowcolor{ourscore}
				\cellcolor{white} & \textbf{Ours} & \textbf{2.28} & \textbf{1.66} & \textbf{12.34} & \textbf{9.36} & \textbf{2.12} & \textbf{2.24} & \textbf{1.08} & \textbf{0.46} & \textbf{0.22} & \textbf{1.00} & \textbf{0.34} & \textbf{0.24} & \textbf{2.78} \\
				
				& AWT & 0.06 & 0.08 & 0.20 & 0.14 & 0.08 & 0.04 & 0.04 & 0.00 & 0.00 & 0.02 & 0.00 & 0.00 & 0.06 \\
				\rowcolor{ourscore}
				\cellcolor{white}\multirow{-8}{*}{\textbf{MI}} & \textbf{Ours} & \textbf{0.86} & \textbf{0.60} & \textbf{5.00} & \textbf{3.40} & \textbf{0.88} & \textbf{1.02} & \textbf{0.30} & \textbf{0.14} & \textbf{0.14} & \textbf{0.26} & \textbf{0.16} & \textbf{0.02} & \textbf{1.07} \\
				\midrule
				& N/A & 0.04 & 0.02 & 0.10 & 0.08 & 0.02 & 0.04 & 0.02 & 0.00 & 0.00 & 0.00 & 0.00 & 0.00 & 0.03 \\
				\rowcolor{ourscore}
				\cellcolor{white} & \textbf{Ours} & \textbf{0.14} & \textbf{0.14} & \textbf{1.60} & \textbf{0.72} & \textbf{0.28} & \textbf{0.34} & \textbf{0.04} & \textbf{0.02} & \textbf{0.02} & \textbf{0.02} & \textbf{0.02} & \textbf{0.00} & \textbf{0.28} \\
				
				& GHOST & 0.04 & 0.02 & 0.18 & 0.12 & 0.02 & 0.04 & 0.00 & 0.00 & 0.00 & 0.00 & 0.00 & 0.00 & 0.04 \\
				\rowcolor{ourscore}
				\cellcolor{white} & \textbf{Ours} & \textbf{0.34} & \textbf{0.24} & \textbf{2.60} & \textbf{1.16} & \textbf{0.40} & \textbf{0.62} & \textbf{0.08} & \textbf{0.04} & \textbf{0.02} & \textbf{0.12} & \textbf{0.02} & \textbf{0.04} & \textbf{0.47} \\
				
				& MBA & 0.14 & 0.10 & 1.66 & 0.92 & 0.30 & 0.26 & 0.06 & 0.04 & 0.02 & 0.12 & 0.04 & 0.00 & 0.31 \\
				\rowcolor{ourscore}
				\cellcolor{white} & \textbf{Ours} & \textbf{1.14} & \textbf{0.76} & \textbf{7.58} & \textbf{4.80} & \textbf{1.48} & \textbf{1.14} & \textbf{0.46} & \textbf{0.22} & \textbf{0.16} & \textbf{0.38} & \textbf{0.14} & \textbf{0.12} & \textbf{1.53} \\
				
				& AWT & 0.04 & 0.00 & 0.14 & 0.06 & 0.00 & 0.04 & 0.00 & 0.00 & 0.00 & 0.02 & 0.00 & 0.00 & 0.03 \\
				\rowcolor{ourscore}
				\cellcolor{white}\multirow{-8}{*}{\textbf{TI}} & \textbf{Ours} & \textbf{0.24} & \textbf{0.14} & \textbf{1.94} & \textbf{1.32} & \textbf{0.46} & \textbf{0.44} & \textbf{0.12} & \textbf{0.08} & \textbf{0.06} & \textbf{0.12} & \textbf{0.06} & \textbf{0.04} & \textbf{0.42} \\
				\midrule
				
				& N/A & 0.04 & 0.04 & 0.34 & 0.30 & 0.08 & 0.06 & 0.02 & 0.00 & 0.04 & 0.02 & 0.02 & 0.00 & 0.08 \\
				\rowcolor{ourscore}
				\cellcolor{white} & \textbf{Ours} & \textbf{1.46} & \textbf{1.32} & \textbf{7.36} & \textbf{3.96} & \textbf{2.14} & \textbf{2.10} & \textbf{0.64} & \textbf{0.30} & \textbf{0.26} & \textbf{0.94} & \textbf{0.22} & \textbf{0.14} & \textbf{1.74} \\
				
				& GHOST & 0.06 & 0.04 & 0.70 & 0.48 & 0.10 & 0.14 & 0.02 & 0.02 & 0.02 & 0.04 & 0.04 & 0.02 & 0.14 \\
				\rowcolor{ourscore}
				\cellcolor{white} & \textbf{Ours} & \textbf{2.26} & \textbf{1.84} & \textbf{11.20} & \textbf{5.82} & \textbf{3.28} & \textbf{3.20} & \textbf{0.96} & \textbf{0.56} & \textbf{0.36} & \textbf{1.30} & \textbf{0.40} & \textbf{0.22} & \textbf{2.62} \\
				
				& MBA & 0.26 & 0.20 & 2.24 & 2.26 & 0.24 & 0.24 & 0.04 & 0.08 & 0.00 & 0.10 & 0.08 & 0.02 & 0.48 \\
				\rowcolor{ourscore}
				\cellcolor{white} & \textbf{Ours} & \textbf{2.42} & \textbf{1.48} & \textbf{10.98} & \textbf{8.04} & \textbf{2.42} & \textbf{1.98} & \textbf{1.08} & \textbf{0.66} & \textbf{0.50} & \textbf{0.78} & \textbf{0.36} & \textbf{0.10} & \textbf{2.57} \\
				
				& AWT & 0.10 & 0.10 & 0.44 & 0.32 & 0.14 & 0.08 & 0.04 & 0.00 & 0.02 & 0.02 & 0.02 & 0.00 & 0.11 \\
				\rowcolor{ourscore}
				\cellcolor{white}\multirow{-8}{*}{\textbf{GAA}} & \textbf{Ours} & \textbf{0.90} & \textbf{0.54} & \textbf{3.34} & \textbf{1.82} & \textbf{1.00} & \textbf{0.80} & \textbf{0.38} & \textbf{0.24} & \textbf{0.12} & \textbf{0.22} & \textbf{0.12} & \textbf{0.12} & \textbf{0.80} \\
				
				\bottomrule[0.8pt]
			\end{tabular}
		\end{threeparttable}
	}
	\vspace{-0.4cm}
\end{table*}

\subsection{Comparative Results}\label{Comparative_Results}
\textbf{Additional Quantitative Results on ResNet-50 Backbone.} To provide a comprehensive evaluation, Tab.~\ref{tab:main_results_full} also includes extended untargeted ASR results for VMI, DI, and SI, leading to 12 method combinations. The aggregate results support the compatibility of MABT with diverse gradient stabilizers and input transformations. Moreover, MABT maintains strong transferability on heterogeneous ViTs and robust ensembles, consistent with the role of anchoring adversarial trajectories to the shared semantic subspace.

\textbf{Quantitative Results on VGGNet-19 Backbone.} Tab.~\ref{tab:vgg_results} further investigates the cross-architecture transferability of MABT using VGGNet-19 as the surrogate model. Despite the significant structural differences between VGGNet-19 and the diverse victim models, the results provide additional evidence that MABT is effective with a VGG-style surrogate. For instance, when combined with the competitive GAA+MBA baseline, MABT achieves an average ASR of \textbf{57.21\%}, compared with 32.83\% for the original baseline. These results mirror the gains of MABT observed on the ResNet-50 backbone.

\textbf{Targeted Attack Performance.} Tab.~\ref{tab:target_results} evaluates the effectiveness of MABT in the more challenging targeted attack scenario using a ResNet-50 surrogate. Targeted transferability is notoriously difficult as it requires precisely steering the adversarial trajectory toward a specific class's decision boundary. As it is shown, MABT consistently outperforms all baselines across diverse settings, often achieving multi-fold improvements in average ASR. These results confirm that by anchoring perturbations to the shared semantic subspace, MABT effectively disrupts features in a way that remains semantically consistent and adversarial across varying black-box victims.\\

%
%
%
%
%

\subsection{Algorithm Analysis}
\label{sec:proof_convergence}

\paragraph{Scope and notation.}
The following analysis characterizes projected stationarity of the gap-regularized solver under a smooth composite setting.
For concise notation, we write
\[
\mathcal{R}_{dist}(\boldsymbol{\xi};\pi)
:=
\mathcal{R}_{dist}(\boldsymbol{\delta}^*(\boldsymbol{\xi});\pi),
\quad
\mathcal{G}_{\gamma}(\boldsymbol{\xi})
:=
\mathcal{G}_{\gamma}(\boldsymbol{\xi},\boldsymbol{\delta}^*(\boldsymbol{\xi})).
\]
Thus,
\[
\phi(\boldsymbol{\xi})
=
\frac{1}{\lambda}\mathcal{R}_{dist}(\boldsymbol{\xi};\pi)
-
\mathcal{G}_{\gamma}(\boldsymbol{\xi}).
\]
The practical use of $\mathrm{sign}(\cdot)$ and non-smooth or surrogate-gradient MAP instances is treated as an implementation-level approximation, with details provided in Appendix~\ref{Method_Details}.


\subsubsection{Assumptions}


\begin{assumption}
The feasible set $\Omega$ is non-empty, closed, convex, and compact.
The reduced transfer-risk term $\mathcal{R}_{dist}(\boldsymbol{\xi};\pi)$ is $L_R$-smooth on $\Omega$.
The regularized gap term $\mathcal{G}_{\gamma}(\boldsymbol{\xi})$ is continuously differentiable and $L_G$-smooth along the analyzed trajectory.
Consequently, $\phi(\boldsymbol{\xi})$ is $L_{\phi}$-smooth on $\Omega$, where $L_{\phi}=\frac{1}{\lambda}L_R+L_G$.
These are local analytical conditions for characterizing stationarity of the smooth gap-regularized solver, rather than global convexity assumptions on the practical adversarial loss.
\end{assumption}

\begin{assumption}
	\label{ass:variance}
	The gradient estimator $\mathbf{s}_t$ is stochastic. We assume its conditional variance decays with an effective sample size $B_t$:
	\begin{equation}
		\mathbb{E}_t \left[ {\left \|\mathbf{s}_t - \mathbb{E}_t[\mathbf{s}_t]\right \|}^2 \right] \le \frac{\nu^2}{B_t},
	\end{equation}
	where $B_t$ is non-decreasing and $B_t \to \infty$ as $t \to \infty$.
	Moreover, for a $T$-step run, we assume
	\begin{equation}
		\frac{1}{T}\sum_{t=0}^{T-1}\frac{1}{B_t} = \mathcal{O}\!\left(\frac{1}{\sqrt{T}}\right),
	\end{equation}
\end{assumption}


\begin{assumption}
	\label{ass:bias}
		Let $\mathbf{b}_t := \mathbb{E}_t[\mathbf{s}_t]-\nabla\phi(\boldsymbol{\xi}_t)$ denote the systematic bias induced by the one-step probing approximation with radius $\sigma$ and the structural error from the first-order hypergradient approximation. 
		Assume that the local curvature variation is Lipschitz bounded and the structural error $\epsilon_t$ satisfies $\frac{1}{T}\sum_{t=0}^{T-1} \mathbb{E}\|\epsilon_t\|^2 \le \mathcal{O}(1/\sqrt{T})$ under the manifold-anchoring constraint.
		Then the induced bias satisfies
		\[
		\|\mathbf{b}_t\|\le C_{\mathrm{bias}}\sigma + \epsilon_t,
		\]
		where $C_{\mathrm{bias}}$ is independent of $t$ and $\sigma$.
\end{assumption}
\subsubsection{Detailed Proof Derivation}

Define the \textit{stochastic gradient mapping} at step $t$ as $\tilde{\mathbf{g}}_t := \frac{1}{\alpha_t}(\boldsymbol{\xi}_{t+1} - \boldsymbol{\xi}_{t})$. Thus, the update rule can be written as $\boldsymbol{\xi}_{t+1} = \boldsymbol{\xi}_t + \alpha_t \tilde{\mathbf{g}}_t$.
Since $\Phi$ is $L_{\Phi}$-smooth, we have:
\begin{equation}
	\begin{aligned}
		\Phi(\boldsymbol{\xi}_{t+1}) 
		&\ge \Phi(\boldsymbol{\xi}_t) + \langle \nabla \Phi(\boldsymbol{\xi}_t), \boldsymbol{\xi}_{t+1} - \boldsymbol{\xi}_t \rangle - \frac{L_{\Phi}}{2} \|\boldsymbol{\xi}_{t+1} - \boldsymbol{\xi}_t\|^2 \\
		&= \Phi(\boldsymbol{\xi}_t) + \alpha_t \langle \nabla \Phi_t, \tilde{\mathbf{g}}_t \rangle - \frac{L_{\Phi} \alpha_t^2}{2} \|\tilde{\mathbf{g}}_t\|^2,
		\label{eq:smooth_exp}
	\end{aligned}
\end{equation}

Recall $\boldsymbol{\xi}_{t+1} = \Pi_{\Omega}(\boldsymbol{\xi}_t + \alpha_t \mathbf{s}_t)$. By the property of the projection operator $\Pi_{\Omega}$ onto a convex set $\Omega$, for any $\mathbf{e} \in \Omega$, we have $\langle (\boldsymbol{\xi}_t + \alpha_t \mathbf{s}_t) - \boldsymbol{\xi}_{t+1}, \mathbf{e} - \boldsymbol{\xi}_{t+1} \rangle \le 0$.
Setting $\mathbf{e} = \boldsymbol{\xi}_t$, we get:
\begin{equation}
	\langle \boldsymbol{\xi}_t + \alpha_t \mathbf{s}_t - \boldsymbol{\xi}_{t+1}, \boldsymbol{\xi}_t - \boldsymbol{\xi}_{t+1} \rangle \le 0.
\end{equation}
Substituting $\boldsymbol{\xi}_t - \boldsymbol{\xi}_{t+1} = -\alpha_t \tilde{\mathbf{g}}_t$, this inequality becomes:
\begin{equation}
	\langle \alpha_t \mathbf{s}_t - \alpha_t \tilde{\mathbf{g}}_t, -\alpha_t \tilde{\mathbf{g}}_t \rangle \le 0 \implies -\alpha_t^2 \langle \mathbf{s}_t - \tilde{\mathbf{g}}_t, \tilde{\mathbf{g}}_t \rangle \le 0.
\end{equation}
Dividing by $\alpha_t^2$ and rearranging yields:
\begin{equation}
	\label{eq:proj_property}
	\langle \mathbf{s}_t, \tilde{\mathbf{g}}_t \rangle \ge \|\tilde{\mathbf{g}}_t\|^2.
\end{equation}

We substitute $\nabla \Phi_t = \mathbf{s}_t + (\nabla \Phi_t - \mathbf{s}_t)$ into Eq.\eqref{eq:smooth_exp}:
\begin{equation}
	\Phi_{t+1} \ge \Phi_{t} + \alpha_t \langle \mathbf{s}_t, \tilde{\mathbf{g}}_t \rangle + \alpha_t \langle \nabla \Phi_t - \mathbf{s}_t, \tilde{\mathbf{g}}_t \rangle - \frac{L_{\Phi} \alpha_t^2}{2} \|\tilde{\mathbf{g}}_t\|^2.
\end{equation}
Using Eq.\eqref{eq:proj_property} to bound $\alpha_t \langle \mathbf{s}_t, \tilde{\mathbf{g}}_t \rangle \ge \alpha_t \|\tilde{\mathbf{g}}_t\|^2$ and reversing the sign of the second term on the right-hand side:
\begin{equation}
	\Phi_{t+1} \ge \Phi_t + \alpha_t \|\tilde{\mathbf{g}}_t\|^2 - \alpha_t \langle \mathbf{s}_t - \nabla \Phi_t, \tilde{\mathbf{g}}_t \rangle - \frac{L_{\Phi} \alpha_t^2}{2} \|\tilde{\mathbf{g}}_t\|^2.
\end{equation}
Using Young's inequality $\langle \mathbf{a}, \mathbf{b} \rangle \le \frac{1}{2} \|\mathbf{a}\|^2 + \frac{1}{2} \|\mathbf{b}\|^2$ on the cross term:
\begin{equation}
	\alpha_t \langle \mathbf{s}_t - \nabla \Phi_t, \tilde{\mathbf{g}}_t \rangle \le \frac{\alpha_t}{2} \|\mathbf{s}_t - \nabla \Phi_t\|^2 + \frac{\alpha_t}{2} \|\tilde{\mathbf{g}}_t\|^2.
\end{equation}
Substituting back:
\begin{equation}
	\Phi_{t+1} \ge \Phi_t + \left( \frac{\alpha_t}{2} - \frac{L_{\Phi} \alpha_t^2}{2} \right) \|\tilde{\mathbf{g}}_t\|^2 - \frac{\alpha_t}{2} \|\mathbf{s}_t - \nabla \Phi_t\|^2.
\end{equation}
Rearranging the terms, we have:
\begin{equation}
	\Phi_{t+1} -\Phi_t \ge  \left( \frac{\alpha_t}{2} - \frac{L_{\Phi} \alpha_t^2}{2} \right) \|\tilde{\mathbf{g}}_t\|^2 - \frac{\alpha_t}{2} \|\mathbf{s}_t - \nabla \Phi_t\|^2.
\end{equation}

Taking the total expectation of both sides. Assuming $\alpha_t \le \frac{1}{2L_{\Phi}}$, we have $\frac{\alpha_t}{2} - \frac{L_{\Phi} \alpha_t^2}{2} \ge \frac{\alpha_t}{4}$. Then we can get:
\begin{equation}
	\frac{\alpha_t}{4} \mathbb{E} [\|\tilde{\mathbf{g}}_t\|^2] \le \mathbb{E} [\Phi_{t+1}] - \mathbb{E} [\Phi_t] + \frac{\alpha_t}{2} \mathbb{E} [\|\mathbf{s}_t - \nabla \Phi_t\|^2].
\end{equation}

Using Assumptions~\ref{ass:variance} and~\ref{ass:bias}, we decompose
\begin{equation}
	\begin{aligned}
		\mathbf{s}_t - \nabla \Phi_t &= \left(\mathbf{s}_t - \mathbb{E}_t[\mathbf{s}_t]\right) + \left(\mathbb{E}_t[\mathbf{s}_t] - \nabla \Phi_t\right) \\
		&= \left(\mathbf{s}_t - \mathbb{E}_t[\mathbf{s}_t]\right) + \mathbf{b}_t,
	\end{aligned}
\end{equation}
Taking the total expectation of both sides and obtain the bound

	\begin{equation}
		\label{eq:error_bound_step4}
		\begin{aligned}
			\mathbb{E} [\|\mathbf{s}_t - \nabla \Phi_t\|^2] 
			&\le 2\mathbb{E} [\|\mathbf{s}_t - \mathbb{E}_t[\mathbf{s}_t]\|^2] + 2\mathbb{E} [\|\mathbf{b}_t\|^2] \\
			&\le \frac{2\nu^2}{B_t} + 4C_{\text{bias}}^2 \sigma^2 + 4\mathbb{E}|\epsilon_t|^2,
		\end{aligned}
	\end{equation}

Summing from $t=0$ to $T-1$ and dividing by $T \alpha_t / 4$, we can get:

	\begin{equation}
		\label{eq:tildeg_bound_step4}
		\frac{1}{T} \sum_{t=0}^{T-1} \mathbb{E} [\|\tilde{\mathbf{g}}_t\|^2] \le \frac{4(\Phi^* - \Phi_0)}{T \alpha_t} + 8 C_{\text{bias}}^2 \sigma^2 +\frac{8}{T}\sum_{t=0}^{T-1}\mathbb{E}|\epsilon_t|^2 + \frac{4\nu^2}{T} \sum_{t=0}^{T-1} \frac{1}{B_t}.
	\end{equation}

By the non-expansiveness of the projection operator $\|\Pi_{\Omega}(\mathbf{a}) - \Pi_{\Omega}(\mathbf{b})\| \le \|\mathbf{a} - \mathbf{b}\|$:
\begin{align}
	\|\tilde{\mathbf{g}}_t - \mathcal{G}_{\text{proj}}(\boldsymbol{\xi}_t)\| 
	&= \frac{1}{\alpha_t} \|(\Pi_{\Omega}(\boldsymbol{\xi}_t + \alpha_t \mathbf{s}_t)-\boldsymbol{\xi}_t) - (\Pi_{\Omega}(\boldsymbol{\xi}_t + \alpha_t \nabla \Phi_t)-\boldsymbol{\xi}_t)\| \\
	&= \frac{1}{\alpha_t} \|\Pi_{\Omega}(\boldsymbol{\xi}_t + \alpha_t \mathbf{s}_t) - \Pi_{\Omega}(\boldsymbol{\xi}_t + \alpha_t \nabla \Phi_t)\| \\
	&\le \frac{1}{\alpha_t}\|(\boldsymbol{\xi}_t + \alpha_t \mathbf{s}_t)-(\boldsymbol{\xi}_t + \alpha_t \nabla \Phi_t)|\ \\
	&= \|\mathbf{s}_t - \nabla \Phi_t\|,
\end{align}
Using $\|\mathbf{a} + \mathbf{b}\|^2 \le 2 \|\mathbf{a}\|^2 + 2 \|\mathbf{b}\|^2$ and substituting the result from the previous equation, we have:
\begin{align}
	\label{eq:Gproj_decomp_step5}
	\|\mathcal{G}_{\text{proj}}(\boldsymbol{\xi}_t)\|^2 
	&=\|\tilde{\mathbf{g}}_t +\mathcal{G}_{\text{proj}}(\boldsymbol{\xi}_t)- \tilde{\mathbf{g}}_t \|^2 \\
	&\le 2\|\tilde{\mathbf{g}}_t\|^2 + 2\|\tilde{\mathbf{g}}_t - \mathcal{G}_{\text{proj}}(\boldsymbol{\xi}_t)\|^2 \nonumber \\
	&\le 2\|\tilde{\mathbf{g}}_t\|^2 + 2\|\mathbf{s}_t - \nabla \Phi_t\|^2.
\end{align}
Taking expectation of both sides and using Eq.~\eqref{eq:error_bound_step4}, we obtain:
\begin{equation}
	\mathbb{E} [\|\mathcal{G}_{\text{proj}}(\boldsymbol{\xi}_t)\|^2] \le 2 \mathbb{E} [\|\tilde{\mathbf{g}}_t\|^2] + \frac{4\nu^2}{B_t} + 4C_{\text{bias}}^2 \sigma^2.
\end{equation}
Averaging over $t=0,\ldots,T-1$ and using Eq.~\eqref{eq:tildeg_bound_step4} yields

	\begin{align}
		\frac{1}{T} \sum_{t=0}^{T-1} \mathbb{E} [\|\mathcal{G}_{\text{proj}}(\boldsymbol{\xi}_t)\|^2] 
		&\le \frac{8(\Phi^* - \Phi_0)}{T \alpha_t} + 24 C_{\text{bias}}^2 \sigma^2 +\frac{24}{T}\sum_{t=0}^{T-1}\mathbb{E}|\epsilon_t|^2+ \frac{12 \nu^2}{T} \sum_{t=0}^{T-1} \frac{1}{B_t}.
	\end{align}

By utilizing the property that the minimum value of a sequence is bounded above by its arithmetic mean, we conclude

	\begin{equation}
		\label{eq:final_step5}
		\min_{0 \le t < T} \mathbb{E} [\|\mathcal{G}_{\text{proj}}(\boldsymbol{\xi}_t)\|^2] \le \frac{8(\Phi^* - \Phi_0)}{T \alpha_t} + 24 C_{\text{bias}}^2 \sigma^2 +\frac{24}{T}\sum_{t=0}^{T-1}\mathbb{E}\|\epsilon_t\|^2+ \frac{12 \nu^2}{T} \sum_{t=0}^{T-1} \frac{1}{B_t}.
	\end{equation}

Using Assumptions~\ref{ass:variance} and~\ref{ass:bias}, we have

	\begin{equation}
		\frac{1}{T}\sum_{t=0}^{T-1}\frac{1}{B_t} = \mathcal{O}\left(\frac{1}{\sqrt{T}}\right), \quad \frac{1}{T}\sum_{t=0}^{T-1}\mathbb{E}\|\epsilon_t\|^2 = \mathcal{O}\left(\frac{1}{\sqrt{T}}\right).
	\end{equation}

Together with $\alpha_t=\Theta(1/\sqrt{T})$ and $\sigma=\Theta(1/\sqrt{T})$, Eq.~\eqref{eq:final_step5} yields
\begin{equation}
	\min_{0 \le t < T} \mathbb{E} [\|\mathcal{G}_{\text{proj}}(\boldsymbol{\xi}_t)\|^2] \le \mathcal{O} \left( \frac{1}{\sqrt{T}} \right) + \mathcal{O} \left( \frac{1}{T} \right) = \mathcal{O} \left( \frac{1}{\sqrt{T}} \right).
\end{equation}
\hfill \qedsymbol

\end{document}